\pdfoutput=1

\documentclass[11pt]{article}

\usepackage[preprint]{acl}

\usepackage{times}
\usepackage{latexsym}

\usepackage[T1]{fontenc}

\usepackage[utf8]{inputenc}

\usepackage{microtype}

\usepackage{inconsolata}

\usepackage[table]{xcolor}
\PassOptionsToPackage{table}{xcolor}
\usepackage{graphicx}  
\usepackage{amsmath}
\usepackage{multirow}
\usepackage{algorithm}
\usepackage{algpseudocode}
\usepackage{hyperref} 
\usepackage{cleveref}
\usepackage{wrapfig}
\usepackage{graphicx}
\usepackage[utf8]{inputenc} 
\usepackage[T1]{fontenc}    
\usepackage{url}            
\usepackage{booktabs}       
\usepackage{amsfonts}       
\usepackage{nicefrac}       
\usepackage{microtype}      
\usepackage{xcolor}         

\title{LaCache: Exact Caching and Precision-Adaptive Inference for Diffusion Large Language Models}

\author{
  Xingru Chen\textsuperscript{1}\thanks{Equal contribution.} \quad
  Zelang Liang\textsuperscript{1}\footnotemark[1] \quad
  Yongjia Ma\textsuperscript{1}\footnotemark[1] \quad
  Jiqing Zhan\textsuperscript{1} \AND 
  Shuling Yang\textsuperscript{1} \quad
  Lian Wen\textsuperscript{1} \quad
  Kun Zhan\textsuperscript{1}\thanks{Corresponding author.} \\[6pt]
  \textsuperscript{1}Li Auto Inc. \\[6pt]
}

\begin{document}
\maketitle

{\let\thefootnote \relax
\footnote{
\hangindent=1.8em
Primary contact \texttt{chenxr@buaa.edu.cn}
}}

\begin{abstract}

Diffusion-based Large Language Models (DLLMs) enable parallel generation via Semi-Autoregressive (SAR) decoding in text generation. However, current methods suffer from severe operator-level redundancy: they recompute the entire sequence during denoising steps, ignoring that the prefix and masked suffix remain invariant within a block. 
We propose \textbf{LaCache}, a training-free acceleration framework that alleviates this redundancy through lossless caching and mixed precision. Specifically,
LaCache employs \textbf{Lossless State Memoization (LSM)} by caching three types of intermediate results: (i) \textit{EmbedCache} for embedding outputs, (ii) \textit{RoPECache} for token-wise pre-attention states, and (iii) \textit{FACache} for the online softmax statistics within FlashAttention. These caches allow the model to skip redundant computation on unchanged tokens without altering the output.                 
To further alleviate memory-bandwidth bottlenecks, LaCache integrates a per-group FP8 quantization strategy for FFN layers, tailored to step-dependent activation distributions across the diffusion process.
Experiments demonstrate that LaCache alone achieves approximately 1.3$\times$ end-to-end speedup over vanilla DLLM. When combined with existing acceleration methods, LaCache reaches up to 40.2$\times$ end-to-end speedup while maintaining comparable task accuracy.




\end{abstract}

\section{Introduction}

\vspace{-2pt}
Large language models (LLMs)~\cite{yang2025qwen3,dubey2024llama,brown2020language,ma2025tuning} are widely used in chat~\cite{mao2024multi}, code generation~\cite{jiang2024survey}, and complex reasoning~\cite{wan2024logicasker,liu2025aligning,feng2026cogportraitfinegrainedeyeregioncontrol}. In deployment, inference latency and compute cost are often the main bottlenecks~\cite{lou2023discrete}. Autoregressive decoding generates tokens sequentially, which limits efficiency and hardware parallelism~\cite{dong2025xgrammar,feng2025ditalker,yang2025qr}. This motivates parallel decoding and non autoregressive generation.
\vspace{-2pt}

\begin{figure}
  \centering
  \includegraphics[width=0.48\textwidth]{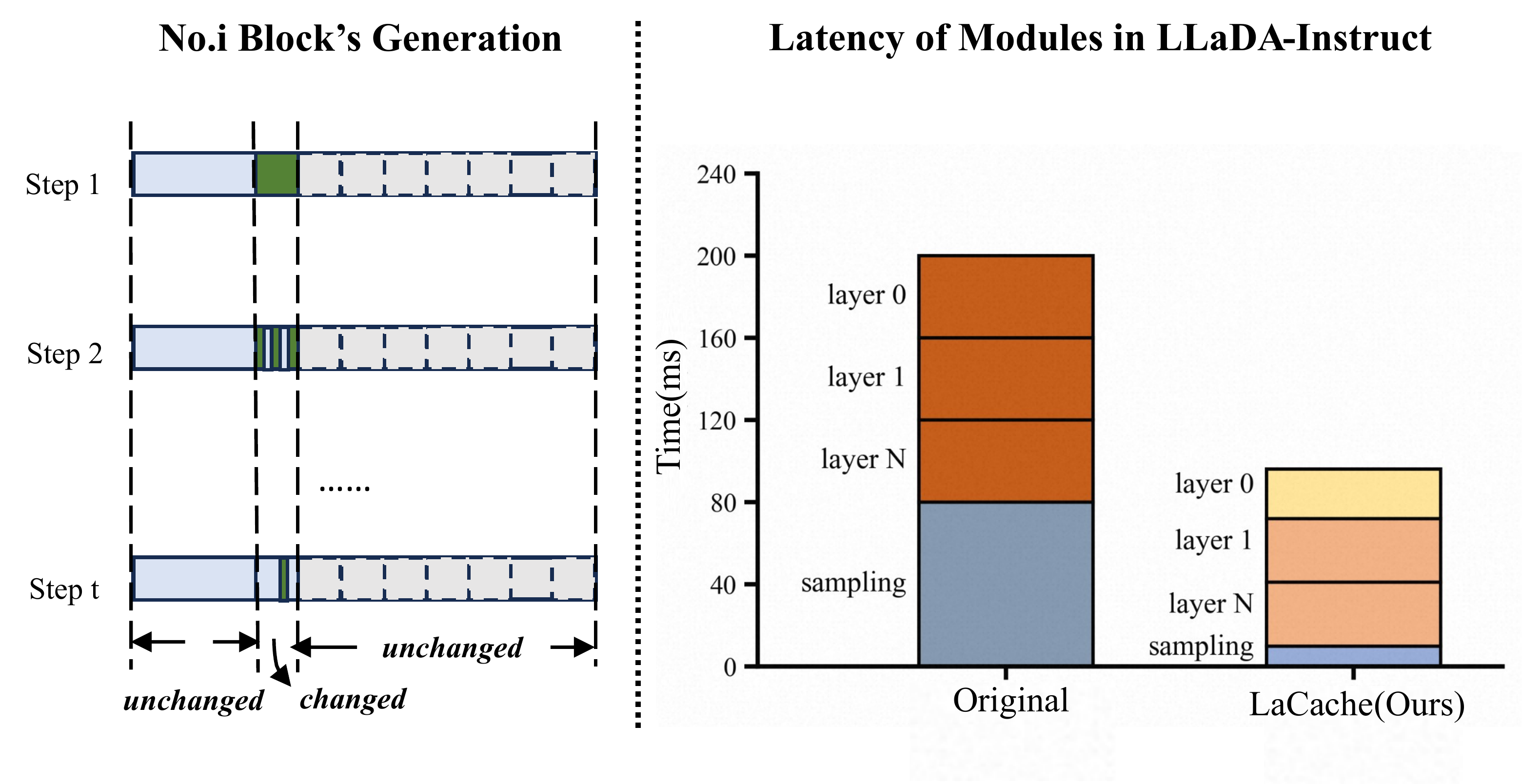}
    \caption{
    \textbf{Left}:DLLMs split total generation tokens into sequential blocks and do parallel decoding across multiple steps inside each block. During one block's generation, the tokens outside of the block stay unchanged, which results in repetition calculation. \textbf{Right}: the latency of the modules in LLaDA-Instruct is mainly present at the transformer layers and the final sampler. With our method LaCache, the latency of the modules is significantly reduced.
    }
    \vspace{-10pt}
    \label{fig:first-pic}
\end{figure}

Diffusion large language models (DLLMs)~\cite{nie2025large,ye2025dream,lyu2019advances, tv3dg_Yang2025,ma2025adams,khanna2025mercury} formulate text generation as iterative denoising, enabling multiple token positions to be updated in one step and employ bidirectional attention to leverage full context. Despite this potential, open-source DLLMs suffer from significant computation overhead~\cite{nie2025large,luo2024trametrajectoryanchoredmultiviewediting}. We attribute this inefficiency to two main constraints: First, bidirectional attention prevents direct reuse of standard KV cache mechanisms. Second, parallel decoding can degrade generation quality due to violations of token dependencies~\cite{wu2025fast,moca_xie2025,rd_nerf_ma2024}.

\begin{figure*}[t]
    \centering
    \includegraphics[width=\textwidth]{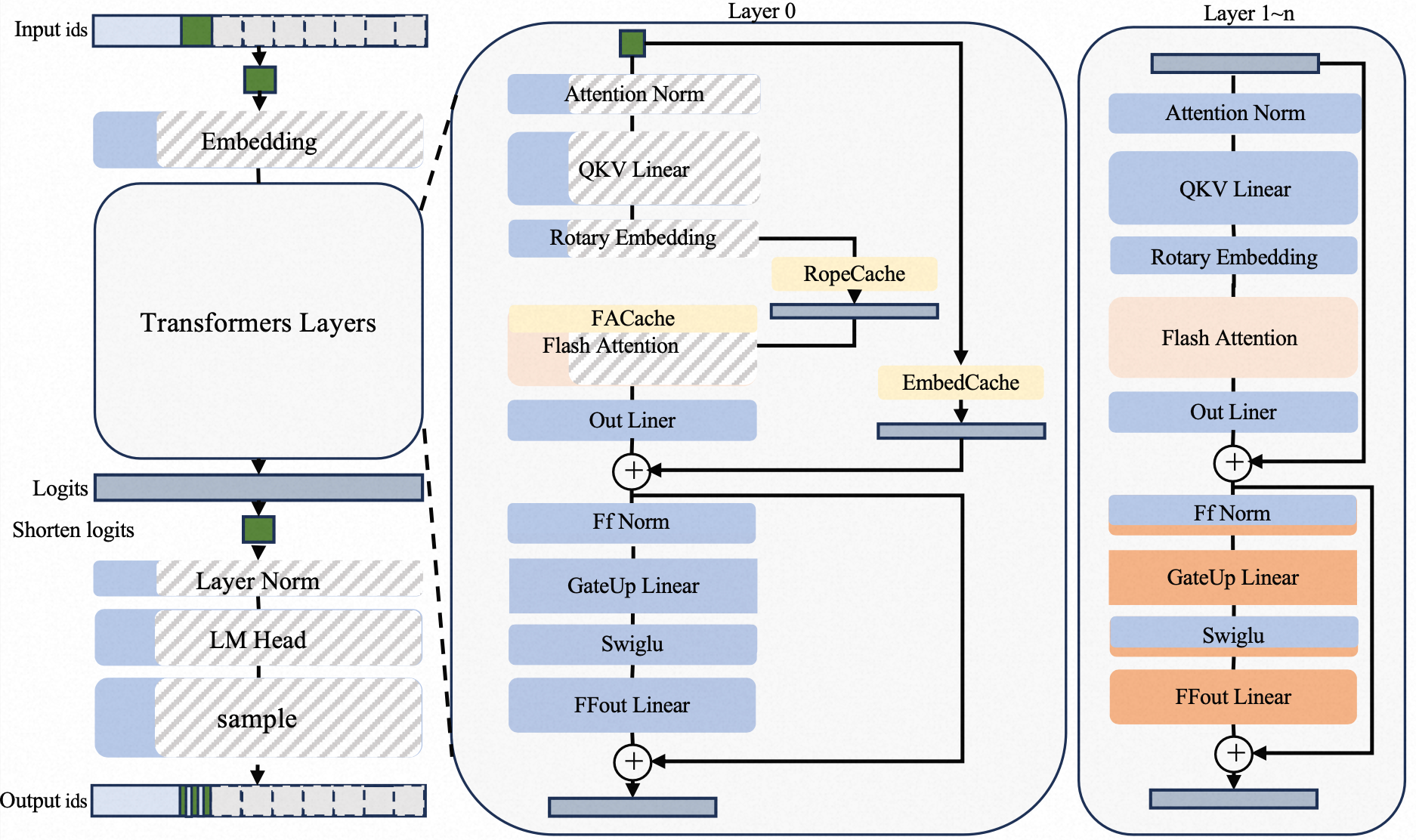}
    \caption{
      The whole procedure of one inference step. After updating the cache in the first step, in the subsequent steps, the input only needs to contain the tokens of the current block, which saves most of the computational work of the embedding layer, norm, qkv linear, rotary embedding and Flash attention in the first layer of transformers. In addition, the method of mixed-precision acceleration is applied to all layers except the first: gateup linear, ffout linear. The FP8 quantization operations are fused with the ff Norm and Swiglu, which further saves overhead. Moreover, The output of the Transformer layers will be truncated again, and only the tokens of the current block will be retained for subsequent sampling operations, which decreased the latency of the float Softmax operation.
    }
     \vspace{-10pt}
    \label{fig:mainstruc}
\end{figure*}

Most open-source DLLMs adopt a semi-autoregressive (SAR)~\cite{nie2024scaling}, block-wise inference scheme. The model initializes future positions with a fixed number of \texttt{[MASK]} tokens and partitions them into blocks. It then performs multiple denoising steps within each block. Across these steps, only the current block tokens change while the prompt and other blocks maintain the same token IDs. Despite this, the model executes a full forward pass at each step, creating substantial redundant computation.
As shown in Fig.~\ref{fig:first-pic} (Left), the high input overlap within a block leads to repeated computation~\cite{wei2025accelerating}. It appears in token wise operators, such as embedding, QKV projections and RoPE. It also appears inside attention computation, especially in the tiled calculation of first layer FlashAttention. The profiling result in Fig.~\ref{fig:first-pic} (right) shows that latency is dominated by Transformer layers and the final sampler. 
This observation suggests that eliminating repeated operator-level computation on unchanged tokens can yield stable and composable end-to-end speedup.
Existing inference acceleration techniques primarily reduce the number of denoising steps or the effective context length. 
Fast DLLM~\cite{wu2025fast} employs confidence threshold decoding and approximate caching to increase throughput. DPad~\cite{chen2025dpad} applies suffix dropout, including a sliding window and distance decay, to reduce the number of tokens used in attention. Other works explore guidance tokens or adaptive decoding schedules~\cite{wei2025orchestrating,sun2025stdd}. 
While these methods are effective and complementary to our approach, they do not directly eliminate the operator-level redundancy caused by unchanged tokens within a SAR block, leaving room for further acceleration.

\vspace{-2pt}


We propose \textbf{LaCache}, a training-free acceleration framework for SAR inference in DLLMs. LaCache reduces operator-level redundancy by reusing intermediate states of unchanged tokens across denoising steps within each SAR block. 
It combines \textbf{Lossless State Memoization (LSM)}, which reuses computation with identical outputs, and a mixed-precision strategy that improves hardware efficiency.
Firstly, \textit{EmbedCache} stores embedding outputs of unchanged tokens. Second, \textit{RoPECache} stores first-layer token-wise pre-attention results, including RoPE. Third, \textit{FACache} stores the online softmax states in first-layer FlashAttention, including the running maximum, the normalizer, and the unnormalized output accumulator, which allow us to skip attention tiles that do not involve the current block. 
Motivated by the diverse activation distributions across layers and denoising steps, LaCache further applies a fine-grained per-group FP8 quantization on FFN layers, which adapts scales to local activation patterns and runs on FP8 Tensor Core kernels fused with surrounding   operators to alleviate memory-bandwidth bottlenecks while preserving model accuracy.
Experiments show that LaCache achieves approximately 1.3$\times$ end-to-end inference speedup over vanilla DLLM. When combined with existing acceleration schemes, it reaches up to 40.2$\times$ end-to-end speedup with comparable quality. Our main contributions are as follows.
\vspace{-2pt}

\begin{itemize}
\vspace{-5pt}
 \item  We identify operator-level redundancy in SAR inference of DLLMs and introduce three lossless cache components that cover token-wise operators and the online softmax state of first-layer FlashAttention, alleviating the redundant computation within each SAR block.
  \vspace{-5pt}
  \item  We integrate a per-group FP8 quantization for FFN layers, adapting to diverse activation distributions and FP8 Tensor Core kernels  to achieve speedup while preserving accuracy.
  \vspace{-5pt}
  \item We demonstrate consistent speedups across multiple models and benchmarks, showing that LaCache composes well with existing acceleration methods.
\end{itemize}

\section{Related Work}

\textbf{Accelerating Diffusion Language Models.}
Recent research optimizes dLLM inference efficiency from multiple perspectives.  
Several works~\cite{jiang2025d,liu2025dllm,wu2025fast} introduce approximate KV cache mechanisms to reuse hidden states across denoising steps. 
Specifically, Fast-dLLM\cite{wu2025fast} enhances throughput via confidence-aware parallel decoding, which replaces top-k sampling with a threshold-based strategy. Dllm-cache ~\cite{liu2025dllm} stored and retrieved key–value (KV) states generated during transformer inference.  From a complementary angle, DPad~\cite{chen2025dpad} reduces the computational cost of attention by pruning the context length, employing suffix dropout with a sliding window.
Other works focus on adaptive decoding schedules, such as inserting guidance tokens~\cite{wei2025orchestrating} or monitoring token stability to dynamically adjust generation thresholds~\cite{sun2025stdd}. While these methods effectively optimize the decoding schedule or context graph, they do not directly eliminate the operator-level redundancy on invariant tokens within a block, 
a gap that our framework is intended to bridge.

\textbf{Low-Precision Inference.}
Quantization has been widely adopted to accelerate LLM inference. SmoothQuant~\cite{xiao2023smoothquant} proposes per-channel quantization that smooths activation outliers across channels to maintain accuracy. COAT~\cite{xi2024coat} introduces mixed-granularity quantization, combining per-tensor and per-group schemes to balance efficiency and precision. Similarly, DeepSeek-V3~\cite{liu2024deepseek} adopts per-group quantization for linear layers. MOSS~\cite{zhang2025moss} refines this with a two-level scaling strategy to reduce precision loss during training. Building on these insights, LaCache applies per-group FP8 quantization to FFN linears in dLLMs, exploiting the activation distribution characteristics observed in SAR inference.

\section{Method}

 \begin{figure}
  \centering
  \vskip -0.2in
    \includegraphics[width=0.48\textwidth]{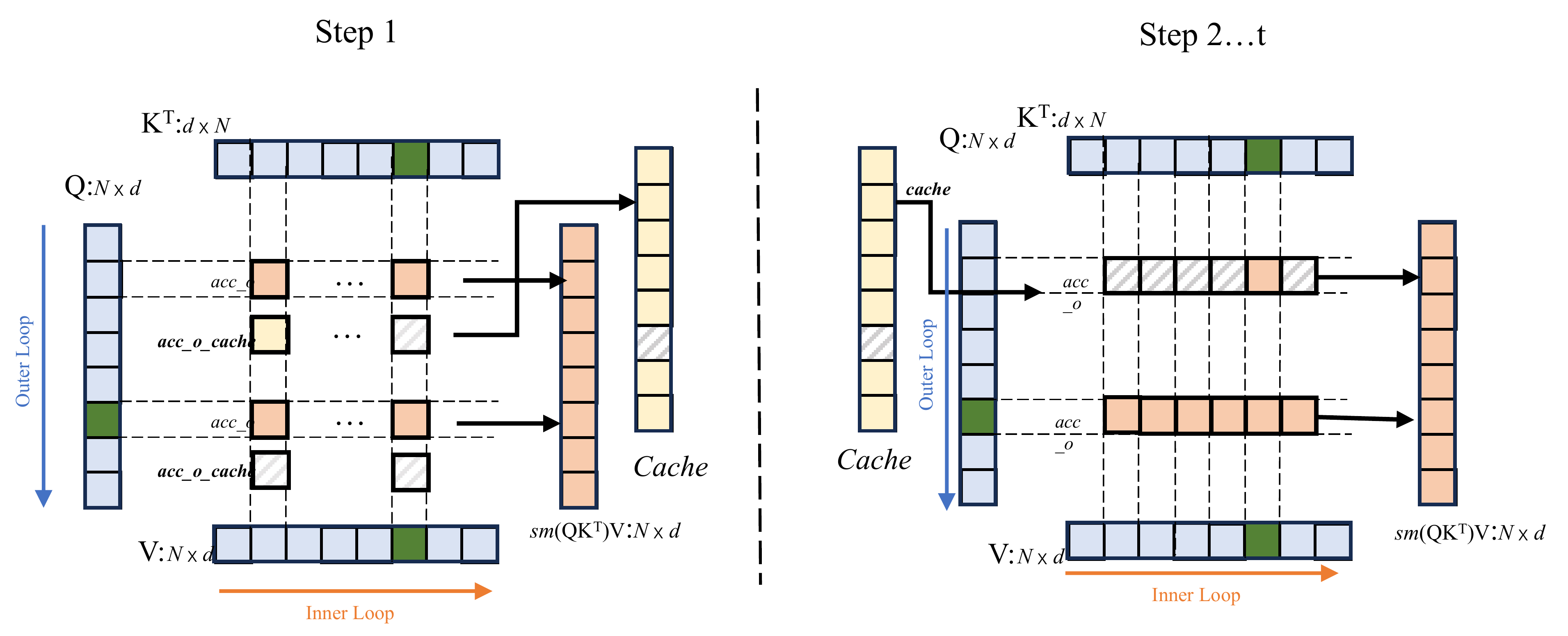}
    \caption{
            The update and resuse scheme of our FACache. \textbf{Left}: Update the cache in the first Flash-attention in the first step.
            When the qkv tile has no overlap with current generation Block, the cache
            follows same calculation procedure to collect the intermidia results.
            \textbf{Right}: Reuse the cache in the first Flash-attention in the following steps.
            When the q tile has no overlap with current generation Block, the registers
            would be initialized with the data in cache. And if the kv tile has no overlap 
            with current generation Block as well, the calculation 
            could be skipped directly.
    }
    \vspace{-10pt}
    \label{fig:attention-cache}
\end{figure}
 
In this section, we start by analyzing the redundant computation in DLLMs, which is raised by the repetition in the input, and the steady activation dynamics suitable for quantization during inference. We then introduce the lossless cache strategy, which contains three parts of the intermediate results during inference. Finally, we propose a fine-grained mixed precision strategy to accelerate the inference with barely precision loss.

\subsection{Overview}
LaCache proposes a Lossless State Memoization (LSM) strategy through the integration of three specialized components: \textit{EmbedCache}, \textit{RoPECache}, and \textit{FACache}. This mechanism ensures mathematically equivalent output while strictly skipping redundant computations for invariant tokens. To further enhance throughput, we incorporate a fine-grained FP8 mixed-precision strategy for compute-intensive linear layers. The overall architecture is presented in Figure~\ref{fig:mainstruc}.


\subsection{The Acceleration space in DLLMs}

\subsubsection{Redundant calculation in DLLM inference}
Within the multiple generation steps of a SAR block, only the tokens in the current block are updated, while the token IDs outside the block remain unchanged. Therefore, redundant computation appears in token-wise operators (embedding, attention norm, QKV projection, RoPE) and in parts of the first-layer FlashAttention tiling procedure. With a proper cache strategy, these redundant costs can be reduced without changing the model output.

\begin{figure}
  \centering
  \includegraphics[width=0.48\textwidth]{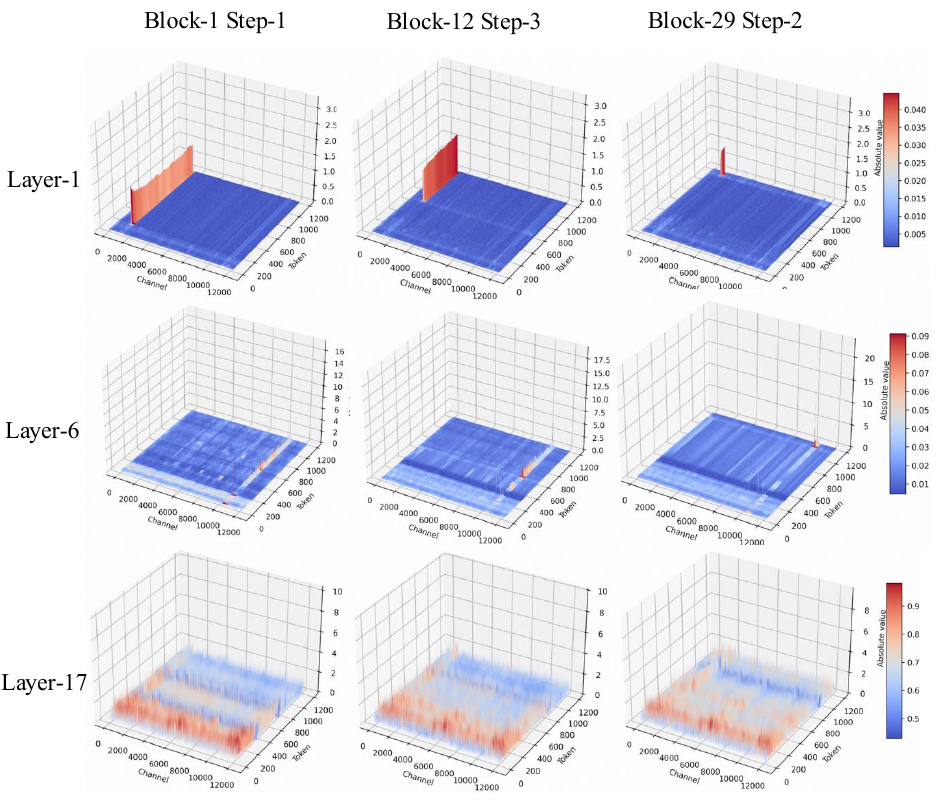}
    \caption{
            The visualization of activations before the ffout linear in LLaDA-Instruct among multiple steps and blocks. 
    }
    \vspace{-10pt}
    \label{fig:activation-vis}
\end{figure}

The structure of DLLMs is still Transformer-based, and linear layers dominate the runtime. A common way to accelerate linear layers is low-precision computation~\cite{kalamkar2019study,lee2009advances,micikevicius2017mixed,liu2024deepseek}. As shown in Fig.~\ref{fig:activation-vis} , the peak of activation in the first layer among multiple steps presents a concentration in several channels, and the peak density decreases as future tokens get unmasked. In middle layers, the activation value is relatively flat with only a few channels having sporadic spikes. In later layers, the values of most channels of the same token are basically the same. This suggest that a fine-grained scheme can better balance efficiency and accuracy, especially under SAR inputs that include many masked tokens.

\subsection{Lossless State Memoization}
To eliminate operator-level redundancy, we introduce Lossless State Memoization (LSM). It strictly memoizes the intermediate results of token-wise operators and the internal online softmax statistics of the first-layer FlashAttention for the invariant region. The memoized states are populated at the first denoising step of each SAR block and reused across subsequent steps within the same block, ensuring mathematically equivalent outputs with reduced computational cost.


\subsubsection{Notation}
Let the full sequence have length $N$. For a SAR block, let $\mathcal{C}$ denote the index set of tokens in the current block (which may change across denoising steps), and let $\mathcal{U} = \{1,\dots,N\}\setminus \mathcal{C}$ denote the remaining tokens (prompt/prefix and masked suffix outside the current block) whose \emph{token IDs} stay unchanged within this block.

\subsubsection{EmbedCache and RoPECache}
Embedding, (RMS)Norm, linear projections, and RoPE are applied independently to each token. Therefore, for any token-wise operator $f$, we have $f(x_{\mathcal{U}})$ unchanged across steps inside the block. We cache:
(i) \textbf{EmbedCache}: the embedding outputs for tokens in $\mathcal{U}$; and
(ii) \textbf{RoPECache}: the first-layer token-wise pre-attention results (after attention norm, QKV projections, and RoPE) for tokens in $\mathcal{U}$.
In subsequent steps, we only compute these token-wise operators for tokens in $\mathcal{C}$ and reuse the cached tensors for $\mathcal{U}$, avoiding redundant work in embedding, attention norm, QKV linear, and RoPE.

Formally, let $H=\mathrm{Embed}(x)$ be the embedding output and let $\hat{H}=\mathrm{Norm}(H)$ be the normalized hidden states at the first layer. The first-layer projections are:
\vspace{-2pt}
\begin{equation}
Q,K = \mathrm{RoPE}(\hat{H}W_{Q,K}),\quad V = \hat{H}W_V.
\end{equation}
Within a SAR block, $x_{\mathcal{U}}$ is unchanged, so $H_{\mathcal{U}}$ and $(Q_{\mathcal{U}},K_{\mathcal{U}},V_{\mathcal{U}})$ can be computed once and reused.

\begin{figure*}[t]
  \vskip 0.2in
  \begin{center}
    \includegraphics[width=\textwidth]{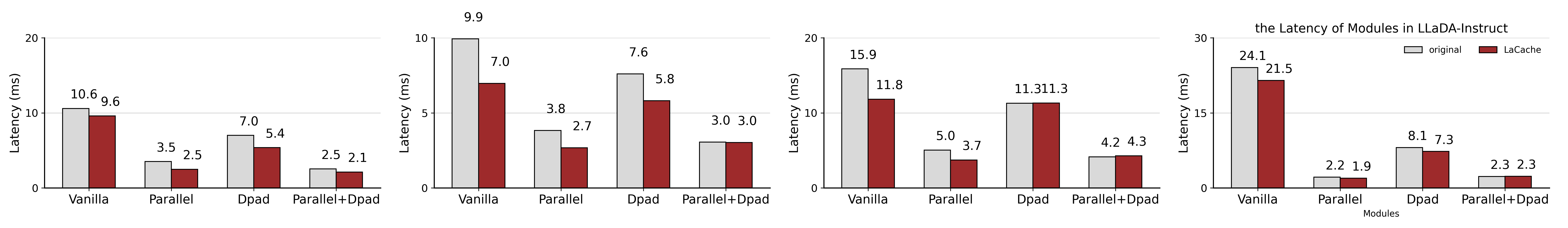}

    \caption{
            The average inference latency of LaCache and previous methods on LLaDA-instruct model among multiple benchmarks. In all tasks, whether combined with parallel or DPad, LaCache can further reduce the latency on the basis of the original method. 
    }
    \vspace{-20pt}
    \label{fig:icml-inference-latency}
  \end{center}
\end{figure*}

\subsubsection{FACache}
\label{sec:fa-state-cache}
The remaining redundancy in the first layer comes from FlashAttention's streaming over key--value blocks. As shown in Fig.~\ref{fig:attention-cache}, for query blocks that belong to $\mathcal{U}$, both the query vectors and the key--value blocks from $\mathcal{U}$ remain unchanged within the SAR block. We therefore cache the \emph{online-softmax state} of FlashAttention for tiles that do not overlap the current block.

Concretely, for a query block $Q_i$ and a streamed KV block $(K_j,V_j)$, define $S_i^{(j)} = Q_i K_j^\top$. FlashAttention maintains per-row state $(m,\ell,\tilde{O})$ and updates it as:
\vspace{-5pt}
\begin{align}
m^{(j)} &= \max\!\left(m^{(j-1)},\ \mathrm{RM}(S^{(j)})\right), \\
\alpha^{(j)} &= \exp\!\left(m^{(j-1)} - m^{(j)}\right), \\
\tilde{P}^{(j)} &= \exp\!\left(S^{(j)} - m^{(j)}\right), \\
\ell^{(j)} &= \alpha^{(j)} \odot \ell^{(j-1)} + \mathrm{RS}(\tilde{P}^{(j)}), \\
\tilde{O}^{(j)} &= \alpha^{(j)} \odot \tilde{O}^{(j-1)} + \tilde{P}^{(j)} V_j,
\end{align}
where $\mathrm{RM}(\cdot)$ and $\mathrm{RS}(\cdot)$ denote the row-wise max and sum respectively. All of the scalings are applied row-wise (with broadcasting over the feature dimension). The final attention output is obtained by row-wise normalization $O=\operatorname{diag}(\ell^{(T_c)})^{-1}\tilde{O}^{(T_c)}$.

Our \textbf{FACache} stores these state variables \emph{restricted to the unchanged region}: for query blocks $Q_i\subseteq \mathcal{U}$, we cache and update $(m_{\text{cache}},\ell_{\text{cache}},\tilde{O}_{\text{cache}})$ to be updated like $({m},\ell,\tilde{O})$ 
after streaming over KV blocks $(K_j,V_j)$ that are fully inside $\mathcal{U}$. 
In implementation, these correspond to the HBM buffers \texttt{RowMaxcache} ($m_{\text{cache}}$), \texttt{RowSumcache} ($\ell_{\text{cache}}$), and \texttt{Ocache} ($\tilde{O}_{\text{cache}}$). 

In subsequent steps within the same SAR block, FlashAttention initializes $(m,\ell,\tilde{O})$ with the cached state and only processes KV blocks that overlap $\mathcal{C}$, while tiles where both query and KV blocks are in $\mathcal{U}$ are skipped entirely.

\subsection{Mixed precision strategy}

As shown in Fig.~\ref{fig:activation-vis}, the peaks of activations are literally not higher than the valleys, and densely exist in masked tokens, which mainly provide position information ~\cite{luxembourg2025plan} instead of semantic information. Moreover, previous methods such as DuQuant ~\cite{lin2024duquant} focus on memory savings without fully exploiting hardware capabilities for acceleration. Therefore we applied the mix-precision strategy in all layers except the first to access acceleration while maintaining precision. The reason to exclude the first layer is that the precision would decrease significantly if applied the quantization in the first layer. Specifically,, We leveraged per-group fp8 quantization before the gate-up linear and ffout linear and applied DeepGEMM in those linears, where activations are less-sensitive, and maintained the original fp16 settings for the left modules. The quantization group on activation is 1x128 while on model weights it’s 128x128. The equation process could be expressed as:
\vspace{-10pt}
\begin{equation}
x'_i = \mathrm{clip}\!\left(\frac{x_i}{scale_g}, Q_{\min}, Q_{\max}\right),\quad i \in g,
\end{equation}
\vspace{-10pt}
\begin{equation}
\text{scale}_g = \frac{\max_{i \in g} |x_i|}{Q_{\max}}.
\end{equation}
Where $x$ is the high precision tensor splitted into $G$ groups, $x'$ is the quantized counterpart, $Q_{\min}$, $Q_{\max}$ are the min and max value of the low-precision format respectively, and $scale_g$ is quantization scale of group $g$.

Meanwhile, to reduce the quantization and dequantization computation overhead, we fused the operation with the AddNorm operation before the gate-up linear, and with the Swiglu operation before ffout linear, which reduced the memory access time~\cite{guo2025sonicmoe}.

Moreover, at the sampling stage, the output would be cut to include only the tokens of current generation block, and therefore accelerate the sampling procedure.


\section{Experiments}

\vspace{-5pt}

\subsection{Experimental setup}

\begin{figure}
  \centering
  \includegraphics[width=0.48\textwidth]{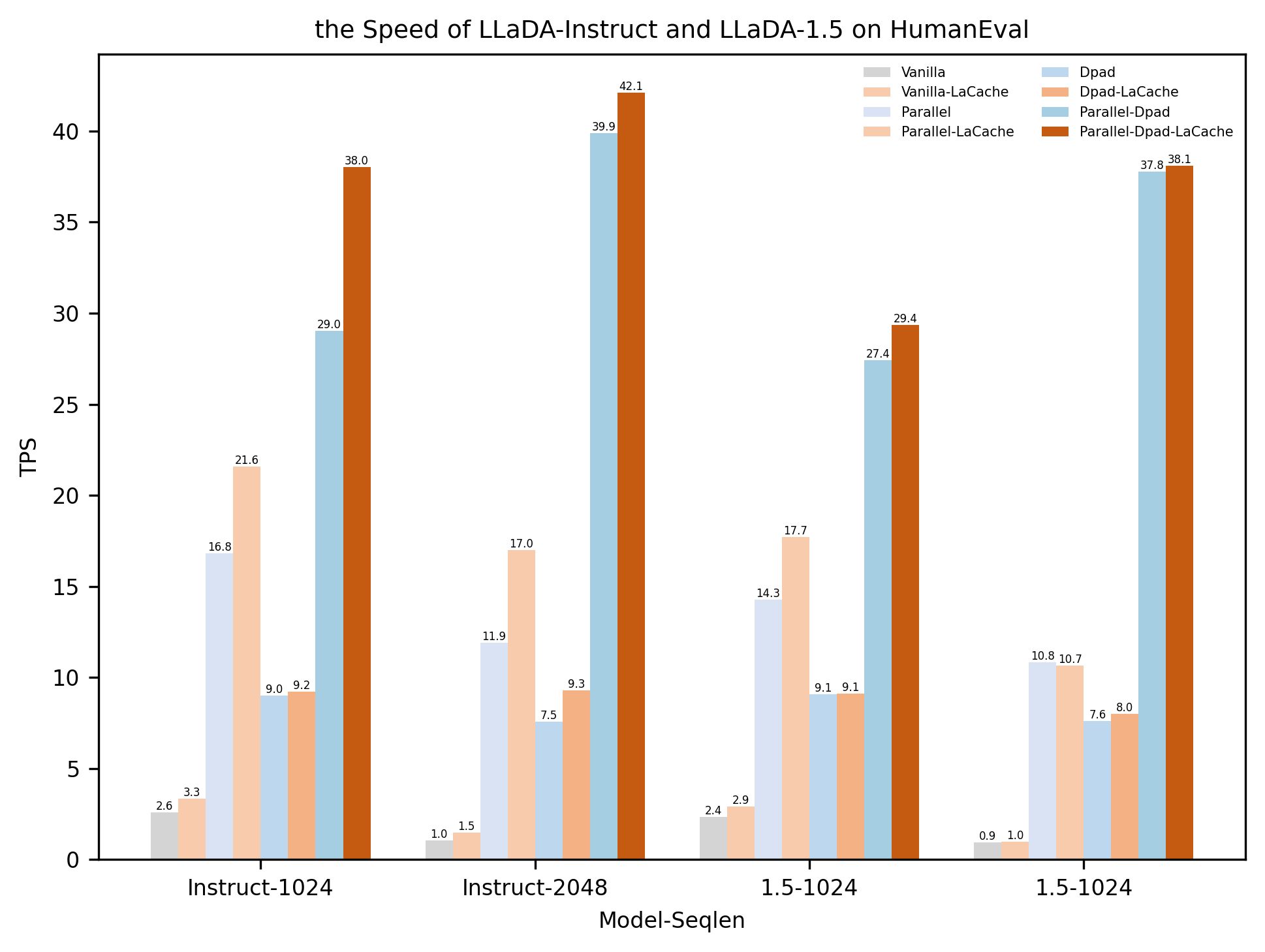}
    \caption{
            Comparison of token generation speed in long text inference scenario of Humaneval benchmark.
    }
    \vspace{-10pt}
    \label{fig:longseq-speed}
\end{figure}

\textbf{Models and Baselines.} All experiments are conducted on an designed GPU. We evaluate LaCache on three variants of diffusion-based large language models: LLaDA(base)~\cite{nie2025large}, LLaDA-instruct and LLaDA-1.5~\cite{zhu2025llada}. Following are the compared baselines:

Vanilla: the original LLaDA~\cite{nie2025large} backbone.

+Parallel(Fast-dllm): a sampling strategy that replaces the top-k strategy in vanilla with a threshold strategy~\cite{wu2025fast} and allows more tokens to be decoded in parallel.

+DPad: a suffix dropout strategy~\cite{chen2025dpad} that drops a number of suffix tokens at each step with a sliding-window or distance-decay strategy.


\textbf{Benchmarks and metrics.} We conduct experiments across a range of task categories. Reasoning performance is evaluated on GSM8K~\cite{cobbe2021training} and MATH~\cite{hendrycks2021measuring}, while code generation is assessed using HumanEval~\cite{chen2021evaluating} and MBPP~\cite{austin2021program}. We examine performance from two complementary perspectives:

\textbf{Accuracy:} Measured using task-dependent evaluation criteria. For code generation, we report pass@1, and for reasoning tasks, we use standard accuracy metrics such as flexible-extract and strict-match accuracy on GSM8K. All accuracy results are presented as percentages.

\textbf{Efficiency:} Assessed in terms of mean inference latency per example and throughput, reported as Tokens Per Second (TPS), which is calculated over the whole output \texttt{<eos>} token is emitted.

\subsection{Main Result}

\begin{table}[ht]

\centering
\footnotesize
    \caption{Accuracy of quantization granularities.}
    \label{tab:compare-quantization_granularites}
    \vspace{-10pt}
    \label{tab:fp8-granularities}
    \small
    \setlength{\tabcolsep}{3pt}
    \renewcommand{\arraystretch}{0.8}
    
    \begin{tabular}{ccccl}
    \cmidrule(lr){1-4}  & 
    \vspace{-10pt} &
    \multicolumn{1}{c}{{\color[HTML]{1F2329} }} \\
{\color[HTML]{1F2329} }                                 & {\color[HTML]{1F2329} }                                   & \multicolumn{2}{c}{{\color[HTML]{1F2329} Accuracy(\%)}} &  \\
\cmidrule(lr){3-4}
\multirow{-2}{*}{{\color[HTML]{1F2329} Benchmark}}      & \multirow{-2}{*}{{\color[HTML]{1F2329} Method}}           & {\color[HTML]{1F2329} Flexible}                      & {\color[HTML]{1F2329} Strict}                        &                                             \\
\cmidrule(lr){1-4}
{\color[HTML]{1F2329} }                                 & {\color[HTML]{1F2329} LaCache}      & {\color[HTML]{1F2329} 76.35} & {\color[HTML]{1F2329} 49.43} &                                             \\
{\color[HTML]{1F2329} }                                 & \cellcolor[HTML]{FFFFFF}{\color[HTML]{1F2329} Per Tensor} & \cellcolor[HTML]{FFFFFF}{\color[HTML]{1F2329} 75.82} & \cellcolor[HTML]{FFFFFF}{\color[HTML]{1F2329} 49.66} &                                             \\
\multirow{-3}{*}{{\color[HTML]{1F2329} GSM8K(4-shot)}}  & \cellcolor[HTML]{FFFFFF}{\color[HTML]{1F2329} Per Token}  & \cellcolor[HTML]{FFFFFF}{\color[HTML]{1F2329} 76.27} & \cellcolor[HTML]{FFFFFF}{\color[HTML]{1F2329} 49.96} &                                             \\
\cmidrule(lr){1-4}
{\color[HTML]{1F2329} }                                 & {\color[HTML]{1F2329} LaCache}      & {\color[HTML]{1F2329} 73.7}  & {\color[HTML]{1F2329} 72.47} &                                             \\
{\color[HTML]{1F2329} }                                 & \cellcolor[HTML]{FFFFFF}{\color[HTML]{1F2329} Per Tensor} & \cellcolor[HTML]{FFFFFF}{\color[HTML]{1F2329} 73.4}  & \cellcolor[HTML]{FFFFFF}{\color[HTML]{1F2329} 72.22} &                                             \\
\multirow{-3}{*}{{\color[HTML]{1F2329} Arc\_easy}}      & \cellcolor[HTML]{FFFFFF}{\color[HTML]{1F2329} Per Token}  & \cellcolor[HTML]{FFFFFF}{\color[HTML]{1F2329} 73.48} & \cellcolor[HTML]{FFFFFF}{\color[HTML]{1F2329} 71.97} &                                             \\
\cmidrule(lr){1-4}
{\color[HTML]{1F2329} }                                 & {\color[HTML]{1F2329} LaCache}      & {\color[HTML]{1F2329} 43.34} & {\color[HTML]{1F2329} 44.97} &                                             \\
{\color[HTML]{1F2329} }                                 & \cellcolor[HTML]{FFFFFF}{\color[HTML]{1F2329} Per Tensor} & \cellcolor[HTML]{FFFFFF}{\color[HTML]{1F2329} 42.83} & \cellcolor[HTML]{FFFFFF}{\color[HTML]{1F2329} 46.33} &                                             \\
\multirow{-3}{*}{{\color[HTML]{1F2329} Arc\_challenge}} & \cellcolor[HTML]{FFFFFF}{\color[HTML]{1F2329} Per Token}  & \cellcolor[HTML]{FFFFFF}{\color[HTML]{1F2329} 42.66} & \cellcolor[HTML]{FFFFFF}{\color[HTML]{1F2329} 46.08} & \\
\cmidrule(lr){1-4}
\end{tabular}

\end{table}

\begin{table}[ht]

\centering
\centering
\footnotesize
    \centering
    \caption{Efficiency of quantization methods}
    \label{tab:compare-quantization_methods}
    \vspace{-10pt}
    \small
    \setlength{\tabcolsep}{3pt}
    \begin{tabular}{ccccc}
    \cmidrule(lr){1-4}  & 
    \vspace{-10pt} &
    \\
{\color[HTML]{1F2329} }                                    & {\color[HTML]{1F2329} }                                         & \multicolumn{2}{c}{{\color[HTML]{1F2329} Efficiency}}                                                                                                             \\
\cmidrule(lr){3-4}
\multirow{-2}{*}{{\color[HTML]{1F2329} Benchmark}}         & \multirow{-2}{*}{{\color[HTML]{1F2329} Method}}                 & {\color[HTML]{1F2329} Latency(s)$\downarrow$}                     & {\color[HTML]{1F2329} TPS$\uparrow$}                                           \\
\cmidrule(lr){1-4}
{\color[HTML]{1F2329} }                                    & {\color[HTML]{1F2329} Vanilla}                                  & {\color[HTML]{1F2329} 10.59}                          & {\color[HTML]{1F2329} 19.78}                        \\
{\color[HTML]{1F2329} }                                    & {\color[HTML]{1F2329} LaCache}            & {\color[HTML]{1F2329} 9.6}    & {\color[HTML]{1F2329} 21.78} \\
\multirow{-3}{*}{{\color[HTML]{1F2329} GSM8K(4-shot)}}     & {\color[HTML]{1F2329} {DuQuant}} & {\color[HTML]{1F2329} 41.18}  & {\color[HTML]{1F2329} 6.22} \\
\cmidrule(lr){1-4}
{\color[HTML]{1F2329} }                                    & {\color[HTML]{1F2329} Vanilla}                                  & {\color[HTML]{1F2329} 9.93}                           & {\color[HTML]{1F2329} 21.11}                        \\
{\color[HTML]{1F2329} }                                    & {\color[HTML]{1F2329} LaCache}            & {\color[HTML]{1F2329} 6.97}   & {\color[HTML]{1F2329} 30.08} \\
\multirow{-3}{*}{{\color[HTML]{1F2329} MATH(4-shot)}}      & {\color[HTML]{1F2329} {DuQuant}} & {\color[HTML]{1F2329} 34.4}   & {\color[HTML]{1F2329} 9.72}\\
\cmidrule(lr){1-4}
{\color[HTML]{1F2329} }                                    & {\color[HTML]{1F2329} Vanilla}                                  & {\color[HTML]{1F2329} 15.87}                          & {\color[HTML]{1F2329} 6.71} \\
{\color[HTML]{1F2329} }                                    & {\color[HTML]{1F2329} LaCache}            & {\color[HTML]{1F2329} 11.83}  & {\color[HTML]{1F2329} 8.73} \\
\multirow{-3}{*}{{\color[HTML]{1F2329} HumanEval(0-shot)}} & {\color[HTML]{1F2329} {DuQuant}} & {\color[HTML]{1F2329} 118.03} & {\color[HTML]{1F2329} 0.86} \\
\cmidrule(lr){1-4}
\end{tabular}
\vspace{-15pt}
\end{table}

\subsubsection{Latency}

Overall, LaCache achieved the fastest speed on all benchmarks and model variants (Tab.~\ref{tab:main-result}) with 10.3\%-42.5\% faster than the Vanilla Backbone and the accuracy largely lossless, even 4 points increased in the HumalEval. When combined with Parallel, it accelerates the backbone by 11.3\%-42.9\%, reaching Vanilla's \textbf{3.7X}-\textbf{12.3X} overall. When combined with DPad, it achieved a 10.5\%-30.8\% acceleration on benchmarks other than humanEval, and a 19\% acceleration on gsm8k when combined with both Parallel and DPad.

In addition, the reason why there is slight acceleration on Humaneval is that the length of prompt is short which results in short length of total input. As shown in Tab.~\ref{tab:long-seq}, we further evaluate with longer generation length 3.7\% speedup over  the results a 3.7\% speedup over  the vanilla backbone, 11.5\% speedup over the parallel backbone, and 8.3\% speedup over the DPad.

Overall, combined with Parallel(fast-DLLM), LaCache can achieve the fastest speed and accuracy without loss, while the speed is slightly inferior when further combined with DPad, probably because the DPad removes more suffix tokens, resulting in shorter gen tokens and less pronounced acceleration.

\begin{figure*}[t]
    \centering
    \includegraphics[width=\textwidth]{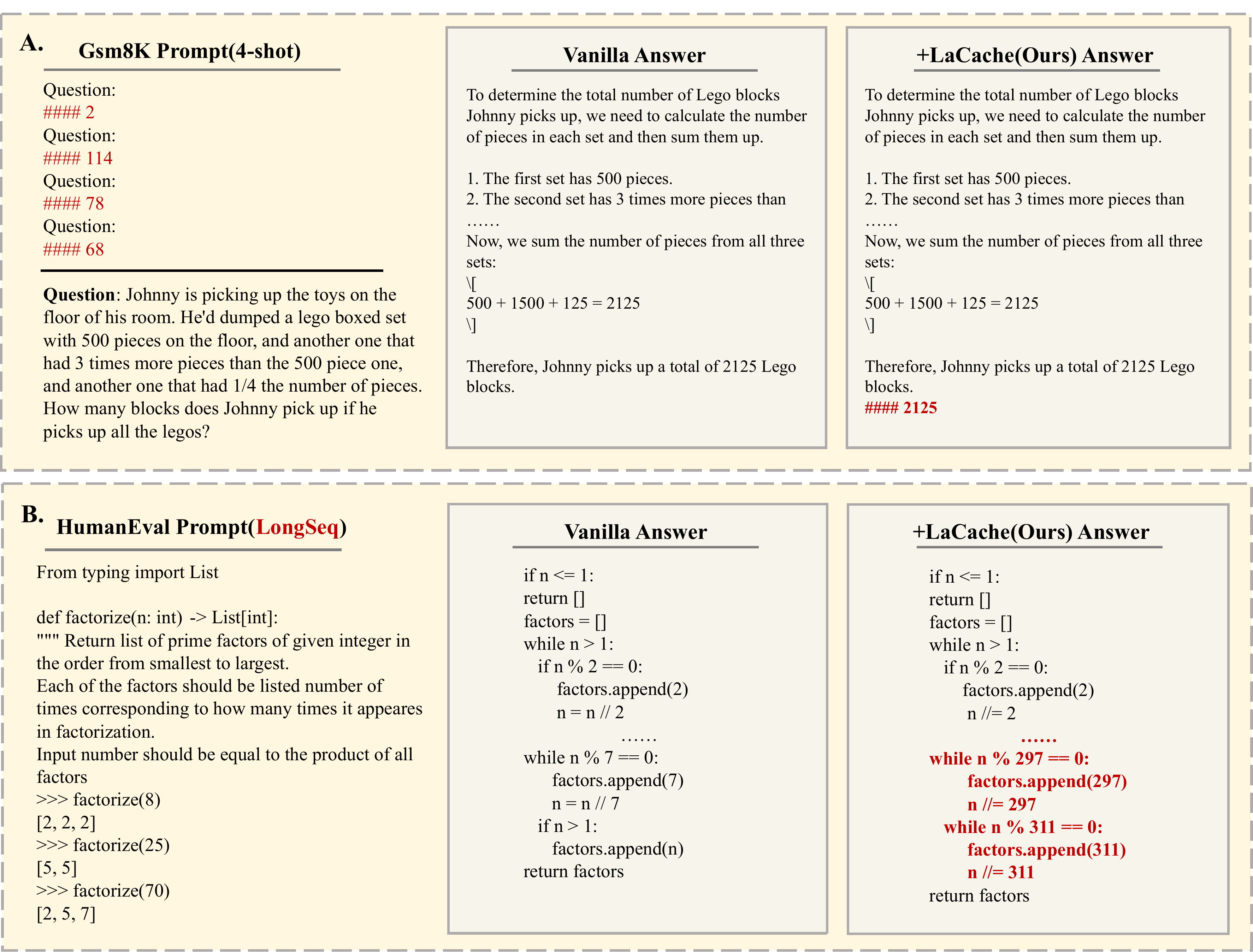}
    \caption{
      The response examples of the backbone and LaCache(Ours) generated on different datasets. A. LaCache's answers are basically the same as backbone in most scenarios, except that the low-precision strategy makes the model slightly tend to generate the end symbol later, so that the correct format of the answer can be output more completely. B. In the scenario of long text generation, when answering difficult code questions using the enumeration method, LaCache tends to generate more enum content, so that the answer is closer to the correct answer.
    }
    \label{fig:answer-sample}
\end{figure*}


\subsubsection{Throughput}

LaCache achieved the best speed in this metric with 10.1\%-42.5\% faster than Vanilla; When combined with parallel, it accelerates by 22.3\%-43.0\%, reaching \textbf{3.7X}-\textbf{12.2X} of Vanilla. When combined with DPad, it accelerates by 12.0\%-31.7\%; When combined with DPad+parallel, it is accelerated by 4.1\%-20.3\%.


\subsubsection{Accuracy}

LaCache has basically no loss of accuracy on all datasets, and there is a slight improvement in the strict-match score. This shows that the low-precision quantization of the per-group does not bring negative losses to the model.

\begin{table}[]
\begin{minipage}[t]{\columnwidth}
\caption{Ablation results of the caching strategy alone.}
\vspace{-10pt}
\label{tab:ablation-cache}
\setlength{\tabcolsep}{4pt}
\footnotesize
\begin{tabular}{ccccl}

\cmidrule(lr){1-4}  & 
\vspace{-10pt} &
\multicolumn{1}{c}{{\color[HTML]{1F2329} }} \\
{\color[HTML]{1F2329} }                                    & {\color[HTML]{1F2329} }                                                  & \multicolumn{2}{c}{{\color[HTML]{1F2329} Efficiency}}                                                    & 
  \\
\cmidrule(lr){3-4}
\multirow{-2}{*}{{\color[HTML]{1F2329} Benchmark}}         & \multirow{-2}{*}{{\color[HTML]{1F2329} Method}}                          & {\color[HTML]{1F2329} Latency(s)}                    & {\color[HTML]{1F2329} TPS}                           &                                             \\
\cmidrule(lr){1-4}
{\color[HTML]{1F2329} }                                    & {\color[HTML]{1F2329} Vanilla}                                           & {\color[HTML]{1F2329} 10.59}                         & {\color[HTML]{1F2329} 19.78}                         &                                             \\
\multirow{-3}{*}{{\color[HTML]{1F2329} GSM8K(4-shot)}}     & {\color[HTML]{1F2329} LaCache\textbf{(w/o fp8)}} & {\color[HTML]{1F2329} 10.54} & {\color[HTML]{1F2329} 19.86} &                                             \\
{\color[HTML]{1F2329} }                                    & {\color[HTML]{1F2329} LaCache}                     & {\color[HTML]{1F2329} 9.6}   & {\color[HTML]{1F2329} 21.78} &                                             \\

\cmidrule(lr){1-4}
{\color[HTML]{1F2329} }                                    & {\color[HTML]{1F2329} Vanilla}                                           & {\color[HTML]{1F2329} 9.93}                          & {\color[HTML]{1F2329} 21.11}                         &                                             \\
\multirow{-3}{*}{{\color[HTML]{1F2329} MATH(4-shot)}}      & {\color[HTML]{1F2329} LaCache\textbf{(w/o fp8)}} & {\color[HTML]{1F2329} 9.09}  & {\color[HTML]{1F2329} 23.04} &                                             \\
{\color[HTML]{1F2329} }                                    & {\color[HTML]{1F2329} LaCache}                     & {\color[HTML]{1F2329} 6.97}  & {\color[HTML]{1F2329} 30.08} &                                             \\

\cmidrule(lr){1-4}
{\color[HTML]{1F2329} }                                    & {\color[HTML]{1F2329} Vanilla}                                           & {\color[HTML]{1F2329} 13.4}                          & {\color[HTML]{1F2329} 7.93}                          &         

\\
\multirow{-3}{*}{{\color[HTML]{1F2329} HumanEval(0-shot)}} & {\color[HTML]{1F2329} LaCache\textbf{(w/o fp8)}} & {\color[HTML]{1F2329} 13}    & {\color[HTML]{1F2329} 8}     &                                             \\
{\color[HTML]{1F2329} }                                    & {\color[HTML]{1F2329} LaCache}                     & {\color[HTML]{1F2329} 12.18} & {\color[HTML]{1F2329} 8.69}  &                                             \\

\cmidrule(lr){1-4}
{\color[HTML]{1F2329} }                                    & {\color[HTML]{1F2329} Vanilla}                                           & {\color[HTML]{1F2329} 24.07}                         & {\color[HTML]{1F2329} 2.68}                          &                                             \\
\multirow{-3}{*}{{\color[HTML]{1F2329} MBPP(3-shot)}}      & {\color[HTML]{1F2329} LaCache\textbf{(w/o fp8)}} & {\color[HTML]{1F2329} 23.9}  & {\color[HTML]{1F2329} 2.7}   &                                      \\
{\color[HTML]{1F2329} }                                    & {\color[HTML]{1F2329} LaCache}                     & {\color[HTML]{1F2329} 21.51} & {\color[HTML]{1F2329} 2.95}  &                                             \\

\cmidrule(lr){1-4}
\end{tabular}
\end{minipage}
\end{table}

\begin{table*}[t]
\vspace{-20pt}
\caption{Comprehensive benchmark results of LLaDA-Instruct with LaCache over four tasks.}

\label{tab:main-result}
\centering
\footnotesize
\renewcommand{\arraystretch}{0.9}
\setlength{\tabcolsep}{4pt}
\resizebox{\textwidth}{!}{
\begin{tabular}{cccccccc}
\hline
 &  & \multicolumn{3}{c}{Efficiency} &  & \multicolumn{2}{c}{Accuracy (\%)} \\
\cmidrule(lr){3-6} \cmidrule(lr){7-8}
Benchmark & Method & Latency (s)$\downarrow$ & TPS$\uparrow$ & Gen. Length &  & Flexible$\uparrow$ & Strict$\uparrow$ \\
\cmidrule(lr){1-6} \cmidrule(lr){7-8}

 & Vanilla & 10.59 & 19.78 & 209/256 &  & 76.04 & 48.67 \\
 & \cellcolor[HTML]{FAF1D1}{+LaCache} & \cellcolor[HTML]{FAF1D1}9.60 & \cellcolor[HTML]{FAF1D1}21.78 & \cellcolor[HTML]{FAF1D1}211/256 & \cellcolor[HTML]{FAF1D1}   & \cellcolor[HTML]{FAF1D1}76.35 & \cellcolor[HTML]{FAF1D1}49.43 \\
 & +Parallel (Fast-dLLM) & 3.52 & 60.44 & 213/256 &  & 78.70 & 50.64 \\
 & \cellcolor[HTML]{FAF1D1}+Parallel + LaCache & \cellcolor[HTML]{FAF1D1}2.48 & \cellcolor[HTML]{FAF1D1}86.33 & \cellcolor[HTML]{FAF1D1}214/256 & \cellcolor[HTML]{FAF1D1} & \cellcolor[HTML]{FAF1D1}77.71 & \cellcolor[HTML]{FAF1D1}51.63 \\
 & +DPad & 7.00 & 20.48 & 143/256 &  & 77.79 & 75.21 \\
 & \cellcolor[HTML]{FAF1D1}+DPad + LaCache & \cellcolor[HTML]{FAF1D1}5.37 & \cellcolor[HTML]{FAF1D1}26.81 & \cellcolor[HTML]{FAF1D1}144/256 & \cellcolor[HTML]{FAF1D1} & \cellcolor[HTML]{FAF1D1}77.18 & \cellcolor[HTML]{FAF1D1}75.36 \\
 & +Parallel + DPad & 2.53 & 56.88 & 144/256 &  & 77.48 & 75.36 \\
\multirow{-8}{*}{GSM8K (4-shot)} & \cellcolor[HTML]{FAF1D1}+Parallel + DPad + LaCache & \cellcolor[HTML]{FAF1D1}\textbf{2.12} & \cellcolor[HTML]{FAF1D1}\textbf{68.43} & \cellcolor[HTML]{FAF1D1}\textbf{145/256} & \cellcolor[HTML]{FAF1D1} & \cellcolor[HTML]{FAF1D1}\textbf{78.92} & \cellcolor[HTML]{FAF1D1}\textbf{76.35} \\
\midrule

 & Vanilla & 9.93 & 21.11 & 210/256 &  & 31.14 & 32.20 \\
 & \cellcolor[HTML]{FAF1D1}+LaCache & \cellcolor[HTML]{FAF1D1}6.97 & \cellcolor[HTML]{FAF1D1}30.08 & \cellcolor[HTML]{FAF1D1}210/256 & \cellcolor[HTML]{FAF1D1} & \cellcolor[HTML]{FAF1D1}30.94 & \cellcolor[HTML]{FAF1D1}32.48 \\
 & +Parallel (Fast-dLLM) & 3.83 & 54.68 & 210/256 &  & 31.08 & 32.18 \\
 & \cellcolor[HTML]{FAF1D1}+Parallel + LaCache & \cellcolor[HTML]{FAF1D1}\textbf{2.68} & \cellcolor[HTML]{FAF1D1}\textbf{78.13} & \cellcolor[HTML]{FAF1D1}210/256 & \cellcolor[HTML]{FAF1D1} & \cellcolor[HTML]{FAF1D1}\textbf{30.84} & \cellcolor[HTML]{FAF1D1}\textbf{32.48} \\
 & +DPad & 7.60 & 24.32 & 185/256 &  & 33.82 & 32.41 \\
 & \cellcolor[HTML]{FAF1D1}+DPad + LaCache & \cellcolor[HTML]{FAF1D1}5.81 & \cellcolor[HTML]{FAF1D1}31.88 & \cellcolor[HTML]{FAF1D1}185/256 &  \cellcolor[HTML]{FAF1D1} & \cellcolor[HTML]{FAF1D1}33.94 & \cellcolor[HTML]{FAF1D1}31.40 \\
 & +Parallel + DPad & 3.05 & 60.70 & 185/256 &  & 34.14 & 31.72 \\
\multirow{-8}{*}{MATH (4-shot)} & \cellcolor[HTML]{FAF1D1}+Parallel + DPad + LaCache & \cellcolor[HTML]{FAF1D1}3.03 & \cellcolor[HTML]{FAF1D1}69.27 & \cellcolor[HTML]{FAF1D1}210/256 & \cellcolor[HTML]{FAF1D1} & \cellcolor[HTML]{FAF1D1}30.94 & \cellcolor[HTML]{FAF1D1}32.36 \\
\midrule

 & Vanilla & 15.87 & 6.71 & 106/512 &  & 35.98 & -- \\
 & \cellcolor[HTML]{FAF1D1}+LaCache & \cellcolor[HTML]{FAF1D1}11.83 & \cellcolor[HTML]{FAF1D1}8.73 & \cellcolor[HTML]{FAF1D1}103/512 & \cellcolor[HTML]{FAF1D1} & \cellcolor[HTML]{FAF1D1}39.63 & \cellcolor[HTML]{FAF1D1}-- \\
 & +Parallel (Fast-dLLM) & 5.04 & 21.22 & 107/512 &  & 35.37 & -- \\
 & \cellcolor[HTML]{FAF1D1}+Parallel + LaCache & \cellcolor[HTML]{FAF1D1}\textbf{3.72} & \cellcolor[HTML]{FAF1D1}\textbf{28.59} & \cellcolor[HTML]{FAF1D1}106/512 & \cellcolor[HTML]{FAF1D1} & \cellcolor[HTML]{FAF1D1}\textbf{39.02} & \cellcolor[HTML]{FAF1D1}-- \\
 & +DPad & 11.29 & 7.89 & 89/512 &  & 39.63 & -- \\
 & \cellcolor[HTML]{FAF1D1}+DPad + LaCache & \cellcolor[HTML]{FAF1D1}11.32 & \cellcolor[HTML]{FAF1D1}8.84 & \cellcolor[HTML]{FAF1D1}100/512 & \cellcolor[HTML]{FAF1D1} & \cellcolor[HTML]{FAF1D1}40.85 & \cellcolor[HTML]{FAF1D1}-- \\
 & +Parallel + DPad & 4.15 & 24.20 & 100/512 &  & 38.11 & -- \\
\multirow{-8}{*}{HumanEval (0-shot)} & \cellcolor[HTML]{FAF1D1}+Parallel + DPad + LaCache & \cellcolor[HTML]{FAF1D1}4.29 & \cellcolor[HTML]{FAF1D1}25.55 & \cellcolor[HTML]{FAF1D1}110/512 & \cellcolor[HTML]{FAF1D1} & \cellcolor[HTML]{FAF1D1}37.80 & \cellcolor[HTML]{FAF1D1}-- \\
\midrule

 & Vanilla & 24.07 & 2.68 & 64/512 &  & 36.80 & -- \\
 & \cellcolor[HTML]{FAF1D1}+LaCache & \cellcolor[HTML]{FAF1D1}21.51 & \cellcolor[HTML]{FAF1D1}2.95 & \cellcolor[HTML]{FAF1D1}64/512 & \cellcolor[HTML]{FAF1D1} & \cellcolor[HTML]{FAF1D1}36.80 & \cellcolor[HTML]{FAF1D1}-- \\
 & +Parallel (Fast-dLLM) & 2.17 & 26.69 & 65/512 &  & 37.40 & -- \\
 & \cellcolor[HTML]{FAF1D1}+Parallel + LaCache & \cellcolor[HTML]{FAF1D1}\textbf{1.95} & \cellcolor[HTML]{FAF1D1}32.65 & \cellcolor[HTML]{FAF1D1}64/512 & \cellcolor[HTML]{FAF1D1}  & \cellcolor[HTML]{FAF1D1}\textbf{37.40} & \cellcolor[HTML]{FAF1D1}-- \\
 & +DPad & 8.07 & 9.51 & 77/512 &  & 36.40 & -- \\
 & \cellcolor[HTML]{FAF1D1}+DPad + LaCache & \cellcolor[HTML]{FAF1D1}7.30 & \cellcolor[HTML]{FAF1D1}12.41 & \cellcolor[HTML]{FAF1D1}91/512 & \cellcolor[HTML]{FAF1D1} & \cellcolor[HTML]{FAF1D1}38.20 & \cellcolor[HTML]{FAF1D1}-- \\
 & +Parallel + DPad & 2.30 & 34.32 & 79/512 &  & 37.60 & -- \\
\multirow{-8}{*}{MBPP (3-shot)} & \cellcolor[HTML]{FAF1D1}+Parallel + DPad + LaCache & \cellcolor[HTML]{FAF1D1}2.33 & \cellcolor[HTML]{FAF1D1}\textbf{35.72} & \cellcolor[HTML]{FAF1D1}\textbf{83/512} & \cellcolor[HTML]{FAF1D1} & \cellcolor[HTML]{FAF1D1}\textbf{37.20} & \cellcolor[HTML]{FAF1D1}-- \\
\bottomrule
\end{tabular}
}
\end{table*}

\subsubsection{Performance on other dllms}

On Llada-base, LaCache is 20\%-30\% faster than the corresponding backbone on GSM8K, MATH, and MBPP datasets, and the final speed reaches \textbf{2.3X-5.9X} of vanilla. On LLaDA-1.5, it is 10\%-20\% faster than the corresponding backbone on GSM8K, MATH, and MBPP datasets, and the final speed reaches \textbf{3.3X-12.3X} of vanilla.

Meanwhile, as shown in Fig.~\ref{fig:answer-sample}, LaCache is able to generate longer enum answers when faced with difficult questions, thus getting closer to the correct answer than backbones.


\subsubsection{Comparison with different quantization methods}


As shown in Tab.~\ref{tab:fp8-granularities} and Tab.~\ref{tab:compare-quantization_methods}, the per-tile fp8 quantization(LaCache) achieves higher accuracy compared with other granularities and higher efficiency than prior methods such as DuQuant~\cite{lin2024duquant}.


\subsubsection{Ablation Study}

We first evaluate the effect of applying caching alone. As shown in Tab.~\ref{tab:ablation-cache}, caching achieves approximately 3\% speedup over the vanilla backbone across all benchmarks, compensating for the inability to apply mixed-precision acceleration to the first layer.

We further compare the accuracy of different quantization granularities. As shown in Tab.~\ref{tab:compare-quantization_granularites}, the accuracy loss grows as the granularity becomes coarser, indicating that per-tile quantization is more suitable for DLLMs than the other two methods.



\section{Conclusion}


We present \textbf{LaCache}, an inference acceleration framework designed to eliminate operator-level redundancy in DLLMs. LaCache introduces \textbf{Lossless State Memoization (LSM)} to reuse invariant intermediate states in token-wise operators and first-layer FlashAttention, complemented by a fine-grained FP8 mixed-precision strategy for compute-bound linear layers. Experiments demonstrate that LaCache consistently achieves lossless acceleration across multiple DLLM architectures and benchmarks. Furthermore, it is fully composable with existing schedule-level optimization methods, delivering 3--12$\times$ speedups on standard sequences and up to 40.2$\times$ on long contexts. LaCache provides a general, training-free, and complementary solution for efficient DLLM deployment without compromising generation quality.



\newpage

\section*{Limitations}

Although the proposed LaCache achieves significant acceleration, certain limitations remain. First,  lossless caching is theoretically restricted to the first layer of the model. Due to the bidirectional propagation of hidden states in subsequent layers, implementing caching mechanisms deeper in the network necessitates approximate methods, which may introduce slight degradations in generation quality. Second, from a hardware perspective, the mixed-precision strategy depends heavily on modern accelerators equipped with native support for low-precision formats, such as FP8. This reliance restricts the direct deployment of the framework on older hardware architectures that lack optimized matrix calculation engines.

\label{sec:bibtex}

\section*{Acknowledgments}

\bibliography{sec/reference}

\newpage

\appendix

\newpage

\section*{Appendix}
\label{sec:appendix}

\section{Impact Statement}

This work proposes a method to improve the efficiency of diffusion large language model (DLLM) inference. By reducing computational costs, our approach can lower the barrier to deploying and experimenting with large-scale models, potentially benefiting research accessibility and real-world applications such as dialogue systems, code generation, and reasoning tasks.

However, like other advances in DLLM efficiency, this work may also contribute to broader adoption of powerful language models, which raises concerns regarding misuse, biased outputs, and the amplification of harmful or misleading content. These risks are not specific to our method but are inherent to the deployment of diffusion large-scale language models in general.

Our work focuses on improving the computational efficiency of existing models and does not introduce new model capabilities or training data. As such, it does not directly increase the risk of misuse beyond what is already present in current DLLM deployments. Responsible use, proper evaluation, and alignment techniques remain essential for mitigating potential negative impacts.

Overall, we believe that the benefits of improved efficiency—such as reduced computational cost and energy consumption—outweigh the potential risks, provided that the technology is applied responsibly.

\section{Preliminary}

\vspace{-5pt}

\subsection{the Inference and Sampling in DLLM}
\vspace{-5pt}

DLLM adopts a semi-autoregressive (SAR) inference paradigm that balances the latency advantages of parallel decoding with the quality and stability of autoregressive generation. Instead of generating one token at a time or decoding the entire sequence in a fully parallel manner, SAR inference produces tokens in blocks, allowing partial parallelism while preserving causal dependencies across blocks.
Given an input prompt $x=(x_1,…,x_{t_0 })$, DLLM generates the output sequence by partitioning the decoding process into blocks of size B. At inference time, the model predicts future tokens simultaneously within each decoding step, yielding candidate tokens $(y_{(t+1)},…,y_{(t+B)})$. This block-wise generation reduces the total number of decoding iterations by a factor of B compared to fully autoregressive decoding.
Importantly, tokens within the same block do not condition on one another; instead, the model relies on a shared contextual representation to predict all tokens jointly. This structure preserves strict causality across blocks while relaxing token-level dependencies within each block, leading to a semi-autoregressive dependency pattern.
\vspace{-5pt}

\subsection{the Foundational Principles of Flash-attention}
\vspace{-5pt}

Self-attention~\cite{vaswani2017attention} has \emph{quadratic} compute and memory cost in the sequence length. FlashAttention reduces the memory overhead by computing attention in tiles and avoiding materializing the full attention score matrix in HBM. Concretely, the sequence dimension $N$ is divided into row blocks of size $B_r$ and column blocks of size $B_c$. Queries are partitioned into $T_r = \lceil N / B_r \rceil$ blocks $Q_1, \ldots, Q_{T_r}$, while keys and values are partitioned into $T_c = \lceil N / B_c \rceil$ blocks $K_1, \ldots, K_{T_c}$ and $V_1, \ldots, V_{T_c}$.
\vspace{-5pt}

For a query block $Q_i$, FlashAttention streams over all key--value blocks $(K_j,V_j)$, computing block scores on-chip:
\vspace{-2pt}
\begin{equation}
S_i^{(j)} = Q_i K_j^\top \in \mathbb{R}^{B_r \times B_c}.
\end{equation}
To compute the softmax without storing all $S_i^{(j)}$, FlashAttention uses an online (log-sum-exp) formulation that maintains per-row \emph{state variables}: a running maximum $m$, a normalizer $\ell$, and an ab-normalized output accumulator $\tilde{O}$. After processing all key--value blocks, the output is obtained by row-wise normalization $O = \tilde{O} / \ell$. We will reuse $(m,\ell,\tilde{O})$ as the cached attention state in Sec.~\ref{sec:fa-state-cache}.

\section{The Speed-Accuracy Trade-off of only Caching Strategy on more layers}

\begin{figure}[h]
  \centering
  \includegraphics[width=0.48\textwidth]{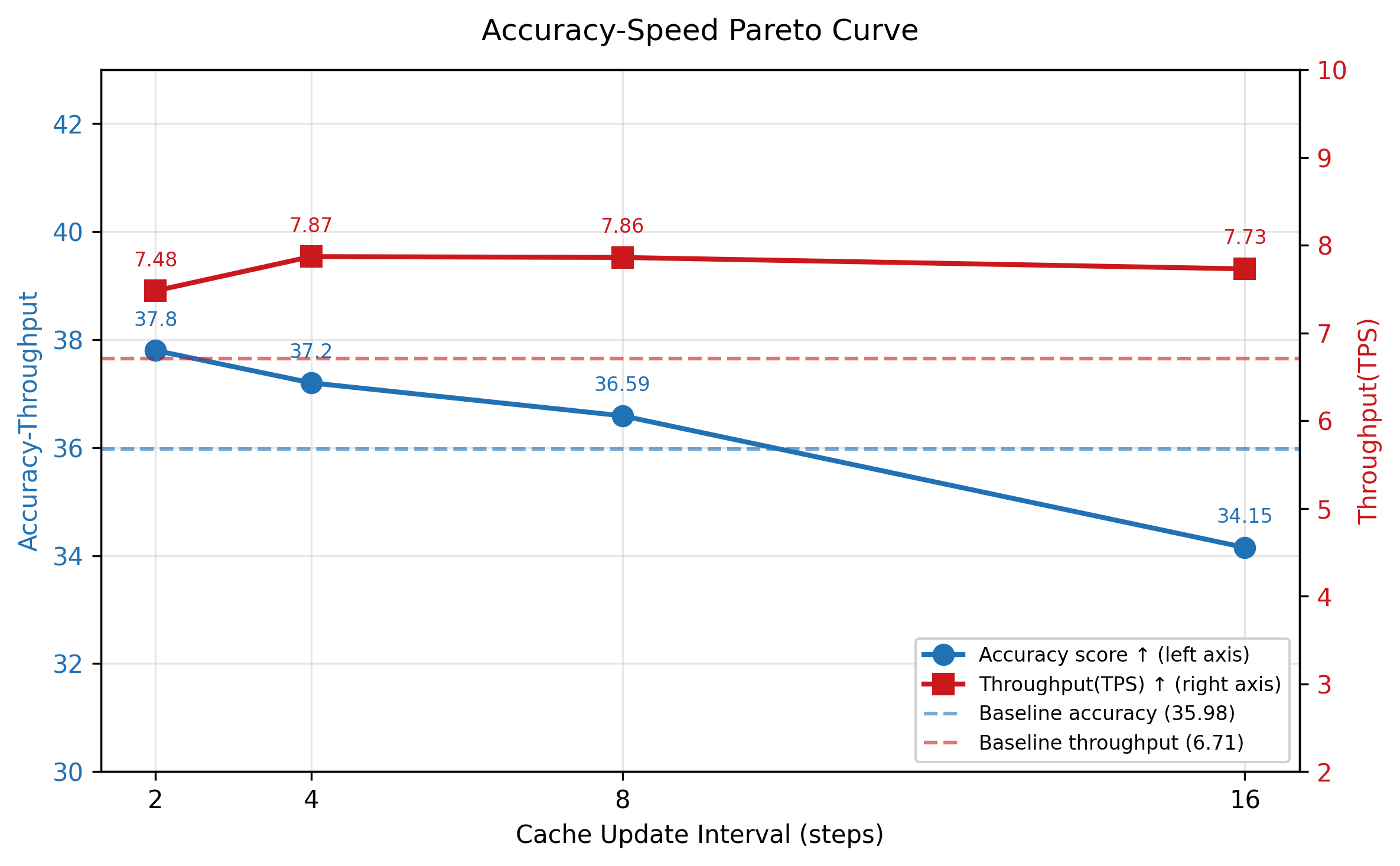}
    \caption{
            The accuracy-speedup curve of caching strategy applied to multiple layers. As shown in the figure, as the cache update interval increases, inference speed improves while accuracy degrades more significantly.
    }
    \label{fig:cache-pareto}
\end{figure}

As shown in Fig.~\ref{fig:cache-pareto}, when applying the caching strategy across multiple layers (layers 0 and 3–17 of LLaDA-Instruct), a speed-accuracy trade-off emerges, governed by the cache update interval. As the interval increases, inference accelerates at the cost of greater accuracy degradation. We empirically identify a configuration that effectively balances efficiency and accuracy.

\section{The acceleration of LaCache on LLaDA-base on multiple benchmarks}

The effect of LaCache on the LLaDA-base model among multiple datasets is presented in Tab.~\ref{tab:base-result}, achieving 10\%-30\% acceleration on datasets other than HumanEval, and the effect is basically lossless. In addition, the reason for the lack of acceleration on the HumanEval dataset is that the overall prompt length of the dataset is short, resulting in insignificant acceleration.

\section{The acceleration of LaCache on LLaDA-1.5 on multiple benchmarks}

The acceleration performance of LaCache on the LLaDA-1.5 model among multiple datasets is presented in Tab.~\ref{tab:1-5-result}, achieving 10\%-30\% acceleration on datasets other than HumanEval, and the effect is basically lossless. In addition, the reason for the lack of acceleration on the HumanEval dataset is that the overall prompt length of the dataset is short, resulting in insignificant acceleration.

\section{The acceleration of LaCache on fast-dllm-v2 on multiple benchmarks}

As shown in Tab.~\ref{tab:fastdllmv2-result}, FACache along could contribute an additional 6\% speedup to Fast-dLLM-V2~\cite{wu2025fast}, demonstrating the complementary acceleration potential of LaCache.

\section{The acceleration of LaCache on LLaDA-Instruct on more benchmarks and more type of GPU}

As shown in Tab.~\ref{tab:more-benchmark} and Tab.~\ref{tab:more-GPU}, LaCache maintains competitive accuracy across a broader range of benchmarks and consistent performance on an another designed GPU.

\begin{table}[t]
\centering
\caption{Additional benchmark results of LLaDA-Instruct with LaCache.}
\footnotesize
\label{tab:more-benchmark}
\begin{tabular}{ccccl}              &  \\
\cmidrule(lr){1-5}
{\color[HTML]{1F2329} }                                        & {\color[HTML]{1F2329} }                         & \multicolumn{2}{c}{{\color[HTML]{1F2329} Accuracy(\%)}}                                                     &  \\
\cmidrule(lr){3-4}

\multirow{-2}{*}{{\color[HTML]{1F2329} Benchmark}}             & \multirow{-2}{*}{{\color[HTML]{1F2329} Method}} & {\color[HTML]{1F2329} Flexible}                      & {\color[HTML]{1F2329} Strict}                        &  \\
\cmidrule(lr){1-5}
{\color[HTML]{1F2329} }                                        & {\color[HTML]{1F2329} Vanilla}                  & {\color[HTML]{1F2329} 73.6}                          & {\color[HTML]{1F2329} 73.3}                          &  \\
\multirow{-2}{*}{{\color[HTML]{1F2329} \textit{piqa(4-shot)}}} & \cellcolor[HTML]{FAF1D1}+LaCache                   & \cellcolor[HTML]{FAF1D1}{\color[HTML]{1F2329} 73.5}  & \cellcolor[HTML]{FAF1D1}{\color[HTML]{1F2329} 73.3}  &  \\
\cmidrule(lr){1-5}
{\color[HTML]{1F2329} }                                        & {\color[HTML]{1F2329} Vanilla}                  & {\color[HTML]{1F2329} 78.7}                          & {\color[HTML]{1F2329} 76.6}                          &  \\
\multirow{-2}{*}{{\color[HTML]{1F2329} arc\_easy}}             & \cellcolor[HTML]{FAF1D1}+LaCache                   & \cellcolor[HTML]{FAF1D1}{\color[HTML]{1F2329} 79}    & \cellcolor[HTML]{FAF1D1}{\color[HTML]{1F2329} 76.2}  &  \\
\cmidrule(lr){1-5}
{\color[HTML]{1F2329} }                                        & {\color[HTML]{1F2329} Vanilla}                  & {\color[HTML]{1F2329} 50}                            & {\color[HTML]{1F2329} 54.5}                          &  \\
\multirow{-2}{*}{{\color[HTML]{1F2329} arc\_chanllenge}}       & \cellcolor[HTML]{FAF1D1}+LaCache                   & \cellcolor[HTML]{FAF1D1}{\color[HTML]{1F2329} 49.53} & \cellcolor[HTML]{FAF1D1}{\color[HTML]{1F2329} 54.6}  &  \\
\cmidrule(lr){1-5}
{\color[HTML]{1F2329} }                                        & {\color[HTML]{1F2329} Vanilla}                  & {\color[HTML]{1F2329} 50.79}                         & {\color[HTML]{1F2329} 67.46}                         &  \\
\multirow{-2}{*}{{\color[HTML]{1F2329} hellaswag}}             & \cellcolor[HTML]{FAF1D1}+LaCache                   & \cellcolor[HTML]{FAF1D1}{\color[HTML]{1F2329} 50.77} & \cellcolor[HTML]{FAF1D1}{\color[HTML]{1F2329} 67.43} &  \\
\cmidrule(lr){1-5}
{\color[HTML]{1F2329} }                                        & {\color[HTML]{1F2329} Vanilla}                  & {\color[HTML]{1F2329} 68.75}                         & {\color[HTML]{1F2329} -}                             &  \\
\multirow{-2}{*}{{\color[HTML]{1F2329} winogrande}}            & \cellcolor[HTML]{FAF1D1}+LaCache                   & \cellcolor[HTML]{FAF1D1}{\color[HTML]{1F2329} 69.06} & \cellcolor[HTML]{FAF1D1}{\color[HTML]{1F2329} -}     &  \\
\cmidrule(lr){1-5}
{\color[HTML]{1F2329} }                                        & {\color[HTML]{1F2329} Vanilla}                  & {27.23} & {27.23} &  \\
\multirow{-2}{*}{{\color[HTML]{1F2329} gpqa}}                  & \cellcolor[HTML]{FAF1D1}+LaCache                   & \cellcolor[HTML]{FAF1D1}{\color[HTML]{1F2329} 30.8}  & \cellcolor[HTML]{FAF1D1}{\color[HTML]{1F2329} 30.8}  & \\
\cmidrule(lr){1-5}
\end{tabular}
\end{table}

\section{Ablation of sequence length}
As shown in Tab.~\ref{tab:long-seq}, the acceleration performance of LaCache on the HumanEval dataset in the scenario of long text generation is presented in Tab.~\Cref{tab:1-5-result}, and shows barely loss.

\section{The answer example of LaCache compared with DPad}

\begin{figure*}[ht]
    \centering
    \vskip -1.55in
    \includegraphics[width=\textwidth]{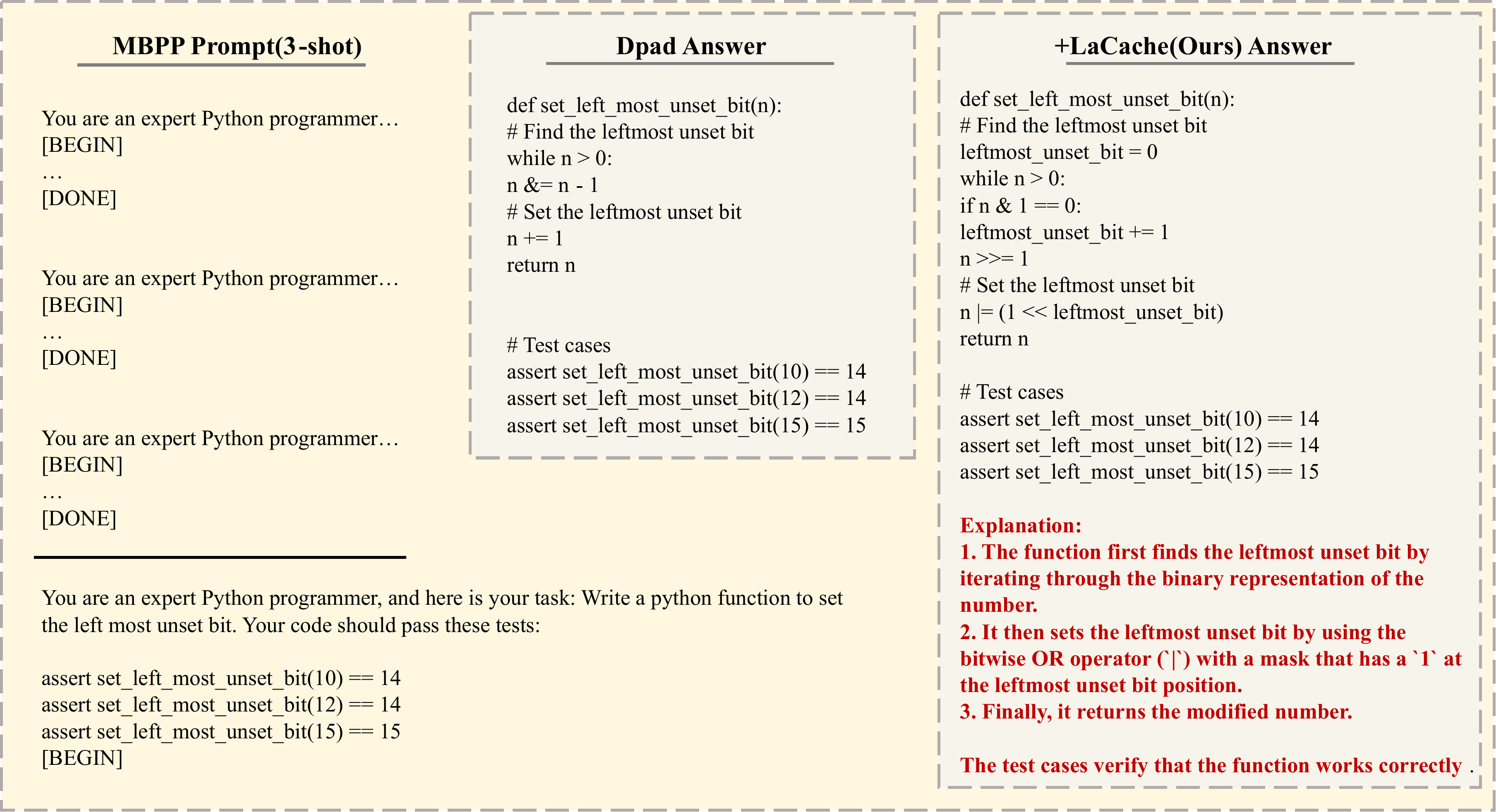}
    \caption{
      The response examples of the DPad backbone and LaCache(Ours) generated with the prompt in MBPP benchmark.
    }
    \label{fig:answer-dpad}
\end{figure*}

As shown in Fig.~\ref{fig:answer-dpad},  although DPad makes the answer shorter, LaCache will make the answer longer, so as to ensure that the comments of the code are fully generated.

\section{The pseudocode of FACache}
As shown below, the FACache is updated in the first generation step and reused in later steps to reduce the redundancy in FlashAttention.

\begin{table*}[ht]
\centering
\caption{Comprehensive benchmark results of LLaDA-base with LaCache over four tasks.}
\label{tab:base-result}
\begin{tabular}{ccccccccc }
\hline
\multicolumn{9}{c}{{\color[HTML]{1F2329} LLaDA-base}}                                                           \\ \hline
{\color[HTML]{1F2329} }                                    & {\color[HTML]{1F2329} }                         & {\color[HTML]{1F2329} }          & \multicolumn{3}{c}{{\color[HTML]{1F2329} Efficiency}}                                      & {\color[HTML]{1F2329} }          & \multicolumn{2}{c}{{\color[HTML]{1F2329} Accuracy(\%)}} \\
\cmidrule(lr){3-6} \cmidrule(lr){7-9}
\multirow{-2}{*}{{\color[HTML]{1F2329} Benchmark}}         & \multirow{-2}{*}{{\color[HTML]{1F2329} Method}} & {\color[HTML]{1F2329} }          & {\color[HTML]{1F2329} Latency(s)}    & {\color[HTML]{1F2329} TPS}            & {\color[HTML]{1F2329} Gen.Length}   & {\color[HTML]{1F2329} }          & {\color[HTML]{1F2329} Flexible}        & {\color[HTML]{1F2329} Strict}          \\
\cmidrule(lr){1-6} \cmidrule(lr){7-9}
{\color[HTML]{1F2329} }                                    & {\color[HTML]{1F2329} Vanilla}                                          & {\color[HTML]{1F2329} }          & {\color[HTML]{1F2329} 10.59}         & {\color[HTML]{1F2329} 14.74}          & {\color[HTML]{1F2329} 156}          & {\color[HTML]{1F2329} }          & {\color[HTML]{1F2329} 62.32}           & {\color[HTML]{1F2329} 69.67}           \\
{\color[HTML]{1F2329} }                                    & {\color[HTML]{1F2329} \cellcolor[HTML]{FAF1D1}+LaCache}                                            & {\color[HTML]{1F2329} \cellcolor[HTML]{FAF1D1}}          & {\color[HTML]{1F2329} \cellcolor[HTML]{FAF1D1}9.66}          & {\color[HTML]{1F2329} \cellcolor[HTML]{FAF1D1}15.82}          & {\color[HTML]{1F2329} \cellcolor[HTML]{FAF1D1}153}          & {\color[HTML]{1F2329}\cellcolor[HTML]{FAF1D1} }          & {\color[HTML]{1F2329} \cellcolor[HTML]{FAF1D1}64.75}           & \cellcolor[HTML]{FAF1D1}{\color[HTML]{1F2329} 69.75}           \\
{\color[HTML]{1F2329} }                                    & {\color[HTML]{1F2329} +Parallel(Fast-dLLM)}                             & {\color[HTML]{1F2329} }          & {\color[HTML]{1F2329} 5.94}          & {\color[HTML]{1F2329} 26.42}          & {\color[HTML]{1F2329} 157}          & {\color[HTML]{1F2329} }          & {\color[HTML]{1F2329} 62.93}           & {\color[HTML]{1F2329} 71.04}           \\
{\color[HTML]{1F2329} }                                    & {\color[HTML]{1F2329} \cellcolor[HTML]{FAF1D1}+Parallel+LaCache}                                   & {\color[HTML]{1F2329}\cellcolor[HTML]{FAF1D1} }          & {\color[HTML]{1F2329} \cellcolor[HTML]{FAF1D1}4.97}          & {\color[HTML]{1F2329} \cellcolor[HTML]{FAF1D1}30.72}          & {\color[HTML]{1F2329} \cellcolor[HTML]{FAF1D1}153}          & {\color[HTML]{1F2329} \cellcolor[HTML]{FAF1D1}}          & {\color[HTML]{1F2329} \cellcolor[HTML]{FAF1D1}65.81}           & {\color[HTML]{1F2329} \cellcolor[HTML]{FAF1D1}70.89}           \\
{\color[HTML]{1F2329} }                                    & {\color[HTML]{1F2329} +DPad}                                            & {\color[HTML]{1F2329} }          & {\color[HTML]{1F2329} 7.47}          & {\color[HTML]{1F2329} 17.25}          & {\color[HTML]{1F2329} 129}          & {\color[HTML]{1F2329} }          & {\color[HTML]{1F2329} 69.83}           & {\color[HTML]{1F2329} 71.57}           \\
{\color[HTML]{1F2329} }                                    & {\color[HTML]{1F2329} \cellcolor[HTML]{FAF1D1}+DPad+LaCache}                                       & {\color[HTML]{1F2329} \cellcolor[HTML]{FAF1D1}}          & {\color[HTML]{1F2329} \cellcolor[HTML]{FAF1D1}6.7}           & {\color[HTML]{1F2329} \cellcolor[HTML]{FAF1D1}19.26}          & {\color[HTML]{1F2329} \cellcolor[HTML]{FAF1D1}129}          & {\color[HTML]{1F2329} \cellcolor[HTML]{FAF1D1}}          & {\color[HTML]{1F2329} \cellcolor[HTML]{FAF1D1}70.66}           & {\color[HTML]{1F2329} \cellcolor[HTML]{FAF1D1}71.04}           \\
{\color[HTML]{1F2329} }                                    & {\color[HTML]{1F2329} +Parallel+DPad}                                   & {\color[HTML]{1F2329} }          & {\color[HTML]{1F2329} 3.85}          & {\color[HTML]{1F2329} 33.57}          & {\color[HTML]{1F2329} 129}          & {\color[HTML]{1F2329} }          & {\color[HTML]{1F2329} 68.92}           & {\color[HTML]{1F2329} 71.04}           \\
\multirow{-8}{*}{{\color[HTML]{1F2329} GSM8K(4-shot)}}     & {\color[HTML]{1F2329} \cellcolor[HTML]{FAF1D1}\textbf{+Parallel+DPad+LaCache}}                     & {\color[HTML]{1F2329} \cellcolor[HTML]{FAF1D1}\textbf{}} & {\color[HTML]{1F2329} \cellcolor[HTML]{FAF1D1}\textbf{3.11}} & {\color[HTML]{1F2329} \cellcolor[HTML]{FAF1D1}\textbf{41.54}} & {\color[HTML]{1F2329} \cellcolor[HTML]{FAF1D1}\textbf{129}} & {\color[HTML]{1F2329} \cellcolor[HTML]{FAF1D1}\textbf{}} & {\color[HTML]{1F2329} \cellcolor[HTML]{FAF1D1}\textbf{69.83}}  & {\color[HTML]{1F2329} \textbf{70.51}}  \\
\midrule
{\color[HTML]{1F2329} }                                    & {\color[HTML]{1F2329} Vanilla}                                          & {\color[HTML]{1F2329} }          & {\color[HTML]{1F2329} 9.14}          & {\color[HTML]{1F2329} 16.69}          & {\color[HTML]{1F2329} 153}          & {\color[HTML]{1F2329} }          & {\color[HTML]{1F2329} 30.1}            & {\color[HTML]{1F2329} 27.54}           \\
{\color[HTML]{1F2329} }                                    & {\color[HTML]{1F2329} \cellcolor[HTML]{FAF1D1}+LaCache}                                            & {\color[HTML]{1F2329}\cellcolor[HTML]{FAF1D1} }          & {\color[HTML]{1F2329} \cellcolor[HTML]{FAF1D1}8.17}          & {\color[HTML]{1F2329}\cellcolor[HTML]{FAF1D1} 18.77}          & {\color[HTML]{1F2329} \cellcolor[HTML]{FAF1D1}153}          & {\color[HTML]{1F2329}\cellcolor[HTML]{FAF1D1} }          & {\color[HTML]{1F2329} \cellcolor[HTML]{FAF1D1}30.48}           & {\color[HTML]{1F2329}\cellcolor[HTML]{FAF1D1} 27.66}           \\
{\color[HTML]{1F2329} }                                    & {\color[HTML]{1F2329} +Parallel(Fast-dLLM)}                             & {\color[HTML]{1F2329} }          & {\color[HTML]{1F2329} 4.98}          & {\color[HTML]{1F2329} 30.67}          & {\color[HTML]{1F2329} 153}          & {\color[HTML]{1F2329} }          & {\color[HTML]{1F2329} 30.2}            & {\color[HTML]{1F2329} 27.66}           \\
{\color[HTML]{1F2329} }                                    & {\color[HTML]{1F2329} \cellcolor[HTML]{FAF1D1}+Parallel+LaCache}                                   & {\color[HTML]{1F2329} \cellcolor[HTML]{FAF1D1}}          & {\color[HTML]{1F2329} \cellcolor[HTML]{FAF1D1}4.1}           & {\color[HTML]{1F2329} \cellcolor[HTML]{FAF1D1}37.47}          & {\color[HTML]{1F2329} \cellcolor[HTML]{FAF1D1}153}          & {\color[HTML]{1F2329}\cellcolor[HTML]{FAF1D1} }          & {\color[HTML]{1F2329} \cellcolor[HTML]{FAF1D1}30.56}           & {\color[HTML]{1F2329} \cellcolor[HTML]{FAF1D1}27.78}           \\
{\color[HTML]{1F2329} }                                    & {\color[HTML]{1F2329} +DPad}                                            & {\color[HTML]{1F2329} }          & {\color[HTML]{1F2329} 8.84}          & {\color[HTML]{1F2329} 17.05}          & {\color[HTML]{1F2329} 151}          & {\color[HTML]{1F2329} }          & {\color[HTML]{1F2329} 30.06}           & {\color[HTML]{1F2329} 27.38}           \\
{\color[HTML]{1F2329} }                                    & {\color[HTML]{1F2329} \cellcolor[HTML]{FAF1D1}+DPad+LaCache}                                       & {\color[HTML]{1F2329}\cellcolor[HTML]{FAF1D1} }          & {\color[HTML]{1F2329} \cellcolor[HTML]{FAF1D1}7.92}          & {\color[HTML]{1F2329} \cellcolor[HTML]{FAF1D1}19.14}          & {\color[HTML]{1F2329} \cellcolor[HTML]{FAF1D1}152}          & {\color[HTML]{1F2329}\cellcolor[HTML]{FAF1D1} }          & {\color[HTML]{1F2329} \cellcolor[HTML]{FAF1D1}30}              & {\color[HTML]{1F2329} \cellcolor[HTML]{FAF1D1}27.52}           \\
{\color[HTML]{1F2329} }                                    & {\color[HTML]{1F2329} +Parallel+DPad}                                   & {\color[HTML]{1F2329} }          & {\color[HTML]{1F2329} 4.74}          & {\color[HTML]{1F2329} 31.8}           & {\color[HTML]{1F2329} 151}          & {\color[HTML]{1F2329} }          & {\color[HTML]{1F2329} 30.1}            & {\color[HTML]{1F2329} 27.42}           \\
\multirow{-8}{*}{{\color[HTML]{1F2329} MATH(4-shot)}}      & {\color[HTML]{1F2329} \cellcolor[HTML]{FAF1D1}\textbf{+Parallel+DPad+LaCache}}                     & {\color[HTML]{1F2329} \cellcolor[HTML]{FAF1D1}\textbf{}} & {\color[HTML]{1F2329} \cellcolor[HTML]{FAF1D1}\textbf{3.91}} & {\color[HTML]{1F2329} \cellcolor[HTML]{FAF1D1}\textbf{38.82}} & {\color[HTML]{1F2329} \cellcolor[HTML]{FAF1D1}\textbf{152}} & {\color[HTML]{1F2329} \cellcolor[HTML]{FAF1D1}\textbf{}} & {\color[HTML]{1F2329} \cellcolor[HTML]{FAF1D1}\textbf{29.94}}  & {\color[HTML]{1F2329} \cellcolor[HTML]{FAF1D1}\textbf{27.54}}  \\
\midrule
{\color[HTML]{1F2329} }                                    & {\color[HTML]{1F2329} Vanilla}                                          & {\color[HTML]{1F2329} }          & {\color[HTML]{1F2329} 13.55}         & {\color[HTML]{1F2329} 8.89}           & {\color[HTML]{1F2329} 120}          & {\color[HTML]{1F2329} }          & {\color[HTML]{1F2329} 33.54}           & {\color[HTML]{1F2329} -}               \\
{\color[HTML]{1F2329} }                                    & {\color[HTML]{1F2329} \cellcolor[HTML]{FAF1D1}+LaCache}                                            & {\color[HTML]{1F2329}\cellcolor[HTML]{FAF1D1} }          & {\color[HTML]{1F2329} \cellcolor[HTML]{FAF1D1}14.45}         & {\color[HTML]{1F2329} \cellcolor[HTML]{FAF1D1}7.44}           & {\color[HTML]{1F2329} \cellcolor[HTML]{FAF1D1}108}          & {\color[HTML]{1F2329}\cellcolor[HTML]{FAF1D1} }          & {\color[HTML]{1F2329} \cellcolor[HTML]{FAF1D1}\cellcolor[HTML]{FAF1D1}32.93}           & {\color[HTML]{1F2329}\cellcolor[HTML]{FAF1D1} -}               \\
{\color[HTML]{1F2329} }                                    & {\color[HTML]{1F2329} +Parallel(Fast-dLLM)}                             & {\color[HTML]{1F2329} }          & {\color[HTML]{1F2329} 4.17}          & {\color[HTML]{1F2329} 28.87}          & {\color[HTML]{1F2329} 120}          & {\color[HTML]{1F2329} }          & {\color[HTML]{1F2329} 33.54}           & {\color[HTML]{1F2329} -}               \\
{\color[HTML]{1F2329} }                                    & {\color[HTML]{1F2329} \cellcolor[HTML]{FAF1D1}+Parallel+LaCache}                                   & {\color[HTML]{1F2329}\cellcolor[HTML]{FAF1D1} }          & {\color[HTML]{1F2329} \cellcolor[HTML]{FAF1D1}3.84}          & {\color[HTML]{1F2329} \cellcolor[HTML]{FAF1D1}28.02}          & {\color[HTML]{1F2329} \cellcolor[HTML]{FAF1D1}108}          & {\color[HTML]{1F2329} \cellcolor[HTML]{FAF1D1}}          & {\color[HTML]{1F2329} \cellcolor[HTML]{FAF1D1}32.93}           & {\color[HTML]{1F2329} \cellcolor[HTML]{FAF1D1}-}               \\
{\color[HTML]{1F2329} }                                    & {\color[HTML]{1F2329} +DPad}                                            & {\color[HTML]{1F2329} }          & {\color[HTML]{1F2329} 9.96}          & {\color[HTML]{1F2329} 9.75}           & {\color[HTML]{1F2329} 97}           & {\color[HTML]{1F2329} }          & {\color[HTML]{1F2329} 32.32}           & {\color[HTML]{1F2329} -}               \\
{\color[HTML]{1F2329} }                                    & {\color[HTML]{1F2329} \cellcolor[HTML]{FAF1D1}+DPad+LaCache}                                       & {\color[HTML]{1F2329}\cellcolor[HTML]{FAF1D1} }          & {\color[HTML]{1F2329} \cellcolor[HTML]{FAF1D1}11.92}         & {\color[HTML]{1F2329} \cellcolor[HTML]{FAF1D1}8.6}            & {\color[HTML]{1F2329} \cellcolor[HTML]{FAF1D1}102}          & {\color[HTML]{1F2329}\cellcolor[HTML]{FAF1D1} }          & {\color[HTML]{1F2329} \cellcolor[HTML]{FAF1D1}34.76}           & {\color[HTML]{1F2329}\cellcolor[HTML]{FAF1D1} -}               \\
{\color[HTML]{1F2329} }                                    & {\color[HTML]{1F2329} +Parallel+DPad}                                   & {\color[HTML]{1F2329} }          & {\color[HTML]{1F2329} 2.36}          & {\color[HTML]{1F2329} 41.12}          & {\color[HTML]{1F2329} 97}           & {\color[HTML]{1F2329} }          & {\color[HTML]{1F2329} 32.32}           & {\color[HTML]{1F2329} -}               \\
\multirow{-8}{*}{{\color[HTML]{1F2329} HumanEval(0-shot)}} & {\color[HTML]{1F2329} \cellcolor[HTML]{FAF1D1}\textbf{+Parallel+DPad+LaCache}}                     & {\color[HTML]{1F2329} \cellcolor[HTML]{FAF1D1}\textbf{}} & {\color[HTML]{1F2329} \cellcolor[HTML]{FAF1D1}\textbf{2.8}}  & {\color[HTML]{1F2329} \cellcolor[HTML]{FAF1D1}\textbf{36.73}} & {\color[HTML]{1F2329} \cellcolor[HTML]{FAF1D1}\textbf{103}} & {\color[HTML]{1F2329} \cellcolor[HTML]{FAF1D1}\textbf{}} & {\color[HTML]{1F2329} \cellcolor[HTML]{FAF1D1}\textbf{35.67}}  & {\color[HTML]{1F2329} \cellcolor[HTML]{FAF1D1}\textbf{-}}      \\
\midrule
{\color[HTML]{1F2329} }                                    & {\color[HTML]{1F2329} Vanilla}                                          & {\color[HTML]{1F2329} }          & {\color[HTML]{1F2329} 23.99}         & {\color[HTML]{1F2329} 2.64}           & {\color[HTML]{1F2329} 63}           & {\color[HTML]{1F2329} }          & {\color[HTML]{1F2329} 40}              & {\color[HTML]{1F2329} -}               \\
{\color[HTML]{1F2329} }                                    & {\color[HTML]{1F2329} \cellcolor[HTML]{FAF1D1}+LaCache}                                            & {\color[HTML]{1F2329} \cellcolor[HTML]{FAF1D1}}          & {\color[HTML]{1F2329} \cellcolor[HTML]{FAF1D1}21.51}         & {\color[HTML]{1F2329} \cellcolor[HTML]{FAF1D1}2.91}           & {\color[HTML]{1F2329} \cellcolor[HTML]{FAF1D1}63}           & {\color[HTML]{1F2329}\cellcolor[HTML]{FAF1D1} }          & {\color[HTML]{1F2329} \cellcolor[HTML]{FAF1D1}39}              & {\color[HTML]{1F2329}\cellcolor[HTML]{FAF1D1} -}               \\
{\color[HTML]{1F2329} }                                    & {\color[HTML]{1F2329} +Parallel(Fast-dLLM)}                             & {\color[HTML]{1F2329} }          & {\color[HTML]{1F2329} 7.32}          & {\color[HTML]{1F2329} 8.63}           & {\color[HTML]{1F2329} 63}           & {\color[HTML]{1F2329} }          & {\color[HTML]{1F2329} 40.4}            & {\color[HTML]{1F2329} -}               \\
{\color[HTML]{1F2329} }                                    & {\color[HTML]{1F2329} \cellcolor[HTML]{FAF1D1}+Parallel+LaCache}                                   & {\color[HTML]{1F2329}\cellcolor[HTML]{FAF1D1} }          & {\color[HTML]{1F2329} \cellcolor[HTML]{FAF1D1}5.96}          & {\color[HTML]{1F2329} \cellcolor[HTML]{FAF1D1}10.5}           & {\color[HTML]{1F2329} \cellcolor[HTML]{FAF1D1}63}           & {\color[HTML]{1F2329} \cellcolor[HTML]{FAF1D1}}          & {\color[HTML]{1F2329} \cellcolor[HTML]{FAF1D1}38.6}            & {\color[HTML]{1F2329} \cellcolor[HTML]{FAF1D1}-}               \\
{\color[HTML]{1F2329} }                                    & {\color[HTML]{1F2329} +DPad}                                            & {\color[HTML]{1F2329} }          & {\color[HTML]{1F2329} 21.11}         & {\color[HTML]{1F2329} 2.85}           & {\color[HTML]{1F2329} 60}           & {\color[HTML]{1F2329} }          & {\color[HTML]{1F2329} 40.6}            & {\color[HTML]{1F2329} -}               \\
{\color[HTML]{1F2329} }                                    & {\color[HTML]{1F2329} \cellcolor[HTML]{FAF1D1}+DPad+LaCache}                                       & {\color[HTML]{1F2329}\cellcolor[HTML]{FAF1D1} }          & {\color[HTML]{1F2329} \cellcolor[HTML]{FAF1D1}18.76}         & {\color[HTML]{1F2329} \cellcolor[HTML]{FAF1D1}3.23}           & {\color[HTML]{1F2329} \cellcolor[HTML]{FAF1D1}61}           & {\color[HTML]{1F2329}\cellcolor[HTML]{FAF1D1} }          & {\color[HTML]{1F2329} \cellcolor[HTML]{FAF1D1}40}              & {\color[HTML]{1F2329}\cellcolor[HTML]{FAF1D1} -}               \\
{\color[HTML]{1F2329} }                                    & {\color[HTML]{1F2329} +Parallel+DPad}                                   & {\color[HTML]{1F2329} }          & {\color[HTML]{1F2329} 4.56}          & {\color[HTML]{1F2329} 13.17}          & {\color[HTML]{1F2329} 60}           & {\color[HTML]{1F2329} }          & {\color[HTML]{1F2329} 40.6}            & {\color[HTML]{1F2329} -}               \\
\multirow{-8}{*}{{\color[HTML]{1F2329} MBPP(3-shot)}}      & {\color[HTML]{1F2329} \cellcolor[HTML]{FAF1D1}\textbf{+Parallel+DPad+LaCache}}                     & {\color[HTML]{1F2329} \cellcolor[HTML]{FAF1D1}\textbf{}} & {\color[HTML]{1F2329} \cellcolor[HTML]{FAF1D1}\textbf{4.09}} & {\color[HTML]{1F2329} \cellcolor[HTML]{FAF1D1}\textbf{14.82}} & {\color[HTML]{1F2329} \cellcolor[HTML]{FAF1D1}\textbf{61}}  & {\color[HTML]{1F2329} \cellcolor[HTML]{FAF1D1}\textbf{}} & {\color[HTML]{1F2329} \cellcolor[HTML]{FAF1D1}\textbf{40}}     & {\color[HTML]{1F2329} \cellcolor[HTML]{FAF1D1}\textbf{-}}     \\
\bottomrule
\end{tabular}
\end{table*}

\begin{table*}[ht]
\centering
\caption{Comprehensive benchmark results of LLaDA-1.5 with LaCache over four tasks.}
\label{tab:1-5-result}
\begin{tabular}{cccccccc}
\hline
\multicolumn{8}{c}{{\color[HTML]{1F2329} LLaDA-1.5}}                                                                                                                                                 \\ \hline
{\color[HTML]{1F2329} }                                    & {\color[HTML]{1F2329} }                         & {\color[HTML]{1F2329} }          & \multicolumn{3}{c}{{\color[HTML]{1F2329} Efficiency}}          & \multicolumn{2}{c}{{\color[HTML]{1F2329} Accuracy(\%)}} \\
\cmidrule(lr){3-6} \cmidrule(lr){7-8}
\multirow{-2}{*}{{\color[HTML]{1F2329} Benchmark}}         & \multirow{-2}{*}{{\color[HTML]{1F2329} Method}} & {\color[HTML]{1F2329} Latency(s)}    & {\color[HTML]{1F2329} TPS}            & {\color[HTML]{1F2329} Gen.Length}   & {\color[HTML]{1F2329} }          & {\color[HTML]{1F2329} Flexible}         & {\color[HTML]{1F2329} Strict}            \\
\cmidrule(lr){1-6} \cmidrule(lr){7-8}
{\color[HTML]{1F2329} }                                    & {\color[HTML]{1F2329} Vanilla}                                          & {\color[HTML]{1F2329} 10.6}          & {\color[HTML]{1F2329} 20.06}          & {\color[HTML]{1F2329} 213}          & {\color[HTML]{1F2329} }          & {\color[HTML]{1F2329} 80.44}            & {\color[HTML]{1F2329} 63.38}                   \\
{\color[HTML]{1F2329} }                                    & {\color[HTML]{1F2329} \cellcolor[HTML]{FAF1D1}{+LaCache}}                                            & \cellcolor[HTML]{FAF1D1}{\color[HTML]{1F2329} 9.65}          & \cellcolor[HTML]{FAF1D1}{22.22}          & \cellcolor[HTML]{FAF1D1}{\color[HTML]{1F2329} 215}          & \cellcolor[HTML]{FAF1D1}{\color[HTML]{1F2329} }          & \cellcolor[HTML]{FAF1D1}{\color[HTML]{1F2329} 79.83}            & \cellcolor[HTML]{FAF1D1}{\color[HTML]{1F2329} 62.47}                    \\
{\color[HTML]{1F2329} }                                    & {\color[HTML]{1F2329} +Parallel(Fast-dLLM)}                             & {\color[HTML]{1F2329} 3.13}          & {\color[HTML]{1F2329} 69.73}          & {\color[HTML]{1F2329} 215}          & {\color[HTML]{1F2329} }          & {\color[HTML]{1F2329} 81.5}             & {\color[HTML]{1F2329} 63.53}                 \\
{\color[HTML]{1F2329} }                                    & {\color[HTML]{1F2329} \cellcolor[HTML]{FAF1D1}+Parallel+LaCache}                                   & {\color[HTML]{1F2329} \cellcolor[HTML]{FAF1D1}2.89}          & {\color[HTML]{1F2329} \cellcolor[HTML]{FAF1D1}75.09}          & {\color[HTML]{1F2329} \cellcolor[HTML]{FAF1D1}217}          & {\color[HTML]{1F2329}\cellcolor[HTML]{FAF1D1} }          & {\color[HTML]{1F2329} \cellcolor[HTML]{FAF1D1}80.21}            & {\color[HTML]{1F2329} \cellcolor[HTML]{FAF1D1}60.8}                \\
{\color[HTML]{1F2329} }                                    & {\color[HTML]{1F2329} +DPad}                                            & {\color[HTML]{1F2329} 6.45}          & {\color[HTML]{1F2329} 22.63}          & {\color[HTML]{1F2329} 146}          & {\color[HTML]{1F2329} }          & {\color[HTML]{1F2329} 80.67}            & {\color[HTML]{1F2329} 80.14}               \\
{\color[HTML]{1F2329} }                                    & {\color[HTML]{1F2329} \cellcolor[HTML]{FAF1D1}+DPad+LaCache}                                       & {\color[HTML]{1F2329} \cellcolor[HTML]{FAF1D1}5.76}          & {\color[HTML]{1F2329} \cellcolor[HTML]{FAF1D1}25.35}          & {\color[HTML]{1F2329} \cellcolor[HTML]{FAF1D1}146}          & {\color[HTML]{1F2329}\cellcolor[HTML]{FAF1D1} }          & {\color[HTML]{1F2329} \cellcolor[HTML]{FAF1D1}81.05}            & {\color[HTML]{1F2329} \cellcolor[HTML]{FAF1D1}79.98}            \\
{\color[HTML]{1F2329} }                                    & {\color[HTML]{1F2329} +Parallel+DPad}                                   & {\color[HTML]{1F2329} 2.26}          & {\color[HTML]{1F2329} 64.55}          & {\color[HTML]{1F2329} 146}          & {\color[HTML]{1F2329} }          & {\color[HTML]{1F2329} 81.8}             & {\color[HTML]{1F2329} 81.05}            \\
\multirow{-8}{*}{{\color[HTML]{1F2329} GSM8K(4-shot)}}     & {\color[HTML]{1F2329} \cellcolor[HTML]{FAF1D1}\textbf{+Parallel+DPad+LaCache}}                     & {\color[HTML]{1F2329} \cellcolor[HTML]{FAF1D1}\textbf{2.05}} & {\color[HTML]{1F2329} \cellcolor[HTML]{FAF1D1}\textbf{71.37}} & {\color[HTML]{1F2329} \cellcolor[HTML]{FAF1D1}\textbf{146}} & {\color[HTML]{1F2329} \cellcolor[HTML]{FAF1D1}} & {\cellcolor[HTML]{FAF1D1}\color[HTML]{1F2329} \cellcolor[HTML]{FAF1D1}\textbf{82.26}}   & {\color[HTML]{1F2329} \cellcolor[HTML]{FAF1D1}\textbf{81.2}} \\
\midrule
{\color[HTML]{1F2329} }                                    & {\color[HTML]{1F2329} Vanilla}                                          & {\color[HTML]{1F2329} 9.09}          & {\color[HTML]{1F2329} 24.01}          & {\color[HTML]{1F2329} 218}          & {\color[HTML]{1F2329} }          & {\color[HTML]{1F2329} 32.9}             & {\color[HTML]{1F2329} 33.66}              \\
{\color[HTML]{1F2329} }                                    & {\color[HTML]{1F2329} \cellcolor[HTML]{FAF1D1}+LaCache}                                            & {\color[HTML]{1F2329} \cellcolor[HTML]{FAF1D1}8.18}          & {\color[HTML]{1F2329} \cellcolor[HTML]{FAF1D1}26.7}           & {\color[HTML]{1F2329} \cellcolor[HTML]{FAF1D1}218}          & {\color[HTML]{1F2329}\cellcolor[HTML]{FAF1D1} }          & {\color[HTML]{1F2329} \cellcolor[HTML]{FAF1D1}33.66}            & {\color[HTML]{1F2329} \cellcolor[HTML]{FAF1D1}34.04}              \\
{\color[HTML]{1F2329} }                                    & {\color[HTML]{1F2329} +Parallel(Fast-dLLM)}                             & {\color[HTML]{1F2329} 3.48}          & {\color[HTML]{1F2329} 62.63}          & {\color[HTML]{1F2329} 218}          & {\color[HTML]{1F2329} }          & {\color[HTML]{1F2329} 32.78}            & {\color[HTML]{1F2329} 33.62}            \\
{\color[HTML]{1F2329} }                                    & {\color[HTML]{1F2329} \cellcolor[HTML]{FAF1D1}+Parallel+LaCache}                                   & {\color[HTML]{1F2329} \cellcolor[HTML]{FAF1D1}3.15}          & {\color[HTML]{1F2329} \cellcolor[HTML]{FAF1D1}69.4}           & {\color[HTML]{1F2329} \cellcolor[HTML]{FAF1D1}218}          & {\color[HTML]{1F2329} \cellcolor[HTML]{FAF1D1}}          & {\color[HTML]{1F2329} \cellcolor[HTML]{FAF1D1}33.58}            & {\color[HTML]{1F2329} \cellcolor[HTML]{FAF1D1}33.82}              \\
{\color[HTML]{1F2329} }                                    & {\color[HTML]{1F2329} +DPad}                                            & {\color[HTML]{1F2329} 6.96}          & {\color[HTML]{1F2329} 27.33}          & {\color[HTML]{1F2329} 190}          & {\color[HTML]{1F2329} }          & {\color[HTML]{1F2329} 36.66}            & {\color[HTML]{1F2329} 33.64}            \\
{\color[HTML]{1F2329} }                                    & {\color[HTML]{1F2329} \cellcolor[HTML]{FAF1D1}+DPad+LaCache}                                       & {\color[HTML]{1F2329} \cellcolor[HTML]{FAF1D1}6.27}          & {\color[HTML]{1F2329} \cellcolor[HTML]{FAF1D1}30.33}          & {\color[HTML]{1F2329} \cellcolor[HTML]{FAF1D1}190}          & {\color[HTML]{1F2329}\cellcolor[HTML]{FAF1D1} }          & {\color[HTML]{1F2329} \cellcolor[HTML]{FAF1D1}36.38}            & {\color[HTML]{1F2329} \cellcolor[HTML]{FAF1D1}33.56}            \\
{\color[HTML]{1F2329} }                                    & {\color[HTML]{1F2329} +Parallel+DPad}                                   & {\color[HTML]{1F2329} 3.04}          & {\color[HTML]{1F2329} 62.38}          & {\color[HTML]{1F2329} 190}          & {\color[HTML]{1F2329} }          & {\color[HTML]{1F2329} 36.04}            & {\color[HTML]{1F2329} 33.24}           \\
\multirow{-8}{*}{{\color[HTML]{1F2329} MATH(4-shot)}}      & {\color[HTML]{1F2329} \cellcolor[HTML]{FAF1D1}\textbf{+Parallel+DPad+LaCache}}                     & {\color[HTML]{1F2329} \cellcolor[HTML]{FAF1D1}\textbf{2.79}} & {\color[HTML]{1F2329} \cellcolor[HTML]{FAF1D1}\textbf{68.37}} & {\color[HTML]{1F2329} \cellcolor[HTML]{FAF1D1}\textbf{191}} & {\color[HTML]{1F2329} \cellcolor[HTML]{FAF1D1}\textbf{}} & {\color[HTML]{1F2329} \cellcolor[HTML]{FAF1D1}\textbf{36.44}}   & {\cellcolor[HTML]{FAF1D1}\color[HTML]{1F2329} \textbf{33.6}}  \\
\midrule
{\color[HTML]{1F2329} }                                    & {\color[HTML]{1F2329} Vanilla}                                          & {\color[HTML]{1F2329} 13.55}         & {\color[HTML]{1F2329} 8}              & {\color[HTML]{1F2329} 108}          & {\color[HTML]{1F2329} }          & {\color[HTML]{1F2329} 41.46}            & {\color[HTML]{1F2329} }               \\
{\color[HTML]{1F2329} }                                    & {\color[HTML]{1F2329} \cellcolor[HTML]{FAF1D1}+LaCache}                                            & {\color[HTML]{1F2329} \cellcolor[HTML]{FAF1D1}14.44}         & {\color[HTML]{1F2329} \cellcolor[HTML]{FAF1D1}7.31}           & {\color[HTML]{1F2329} \cellcolor[HTML]{FAF1D1}106}          & {\color[HTML]{1F2329}\cellcolor[HTML]{FAF1D1} }          & {\color[HTML]{1F2329}\cellcolor[HTML]{FAF1D1} 42.07}            & {\color[HTML]{1F2329} \cellcolor[HTML]{FAF1D1}}               \\
{\color[HTML]{1F2329} }                                    & {\color[HTML]{1F2329} +Parallel(Fast-dLLM)}                             & {\color[HTML]{1F2329} 4.41}          & {\color[HTML]{1F2329} 24.48}          & {\color[HTML]{1F2329} 108}          & {\color[HTML]{1F2329} }          & {\color[HTML]{1F2329} 39.63}            & {\color[HTML]{1F2329} }               \\
{\color[HTML]{1F2329} }                                    & {\color[HTML]{1F2329} \cellcolor[HTML]{FAF1D1}+Parallel+LaCache}                                   & {\color[HTML]{1F2329} \cellcolor[HTML]{FAF1D1}4.58}          & {\color[HTML]{1F2329} \cellcolor[HTML]{FAF1D1}23.47}          & {\color[HTML]{1F2329} \cellcolor[HTML]{FAF1D1}107}          & {\color[HTML]{1F2329}\cellcolor[HTML]{FAF1D1} }          & {\color[HTML]{1F2329} \cellcolor[HTML]{FAF1D1}41.46}            & {\color[HTML]{1F2329}\cellcolor[HTML]{FAF1D1} }               \\
{\color[HTML]{1F2329} }                                    & {\color[HTML]{1F2329} +DPad}                                            & {\color[HTML]{1F2329} 10.07}         & {\color[HTML]{1F2329} 9.66}           & {\color[HTML]{1F2329} 97}           & {\color[HTML]{1F2329} }          & {\color[HTML]{1F2329} 39.63}            & {\color[HTML]{1F2329} }            \\
{\color[HTML]{1F2329} }                                    & {\color[HTML]{1F2329} \cellcolor[HTML]{FAF1D1}+DPad+LaCache}                                       & {\color[HTML]{1F2329} \cellcolor[HTML]{FAF1D1}11.84}         & {\color[HTML]{1F2329} \cellcolor[HTML]{FAF1D1}7.79}           & {\color[HTML]{1F2329} \cellcolor[HTML]{FAF1D1}92}           & {\color[HTML]{1F2329}\cellcolor[HTML]{FAF1D1} }          & {\color[HTML]{1F2329} \cellcolor[HTML]{FAF1D1}39.94}            & {\color[HTML]{1F2329}\cellcolor[HTML]{FAF1D1} }           \\
{\color[HTML]{1F2329} }                                    & {\color[HTML]{1F2329} +Parallel+DPad}                                   & {\color[HTML]{1F2329} 3.9}           & {\color[HTML]{1F2329} 24.28}          & {\color[HTML]{1F2329} 95}           & {\color[HTML]{1F2329} }          & {\color[HTML]{1F2329} 41.46}            & {\color[HTML]{1F2329} }             \\
\multirow{-8}{*}{{\color[HTML]{1F2329} HumanEval(0-shot)}} & {\color[HTML]{1F2329} \cellcolor[HTML]{FAF1D1}\textbf{+Parallel+DPad+LaCache}}                     & {\color[HTML]{1F2329} \cellcolor[HTML]{FAF1D1}\textbf{4.6}}  & {\color[HTML]{1F2329} \cellcolor[HTML]{FAF1D1}\textbf{19.81}} & {\color[HTML]{1F2329} \cellcolor[HTML]{FAF1D1}\textbf{91}}  & {\color[HTML]{1F2329} \cellcolor[HTML]{FAF1D1}\textbf{}} & {\color[HTML]{1F2329} \cellcolor[HTML]{FAF1D1}\textbf{40.55}}   & {\color[HTML]{1F2329} \cellcolor[HTML]{FAF1D1}\textbf{}}     \\
\midrule
{\color[HTML]{1F2329} }                                    & {\color[HTML]{1F2329} Vanilla}                                          & {\color[HTML]{1F2329} 24.07}         & {\color[HTML]{1F2329} 2.63}           & {\color[HTML]{1F2329} 63}           & {\color[HTML]{1F2329} }          & {\color[HTML]{1F2329} 38.8}             & {\color[HTML]{1F2329} }             \\
{\color[HTML]{1F2329} }                                    & {\color[HTML]{1F2329} \cellcolor[HTML]{FAF1D1}+LaCache}                                            & {\color[HTML]{1F2329} \cellcolor[HTML]{FAF1D1}21.48}         & {\color[HTML]{1F2329} \cellcolor[HTML]{FAF1D1}3.11}           & {\color[HTML]{1F2329} \cellcolor[HTML]{FAF1D1}\cellcolor[HTML]{FAF1D1}67}           & {\color[HTML]{1F2329}\cellcolor[HTML]{FAF1D1} }          & {\color[HTML]{1F2329} \cellcolor[HTML]{FAF1D1}38.2}             & {\color[HTML]{1F2329}\cellcolor[HTML]{FAF1D1} }       \\
{\color[HTML]{1F2329} }                                    & {\color[HTML]{1F2329} +Parallel(Fast-dLLM)}                             & {\color[HTML]{1F2329} 2.15}          & {\color[HTML]{1F2329} 29.43}          & {\color[HTML]{1F2329} 63}           & {\color[HTML]{1F2329} }          & {\color[HTML]{1F2329} 39.4}             & {\color[HTML]{1F2329} }         \\
{\color[HTML]{1F2329} }                                    & {\color[HTML]{1F2329} \cellcolor[HTML]{FAF1D1}+Parallel+LaCache}                                   & {\color[HTML]{1F2329}\cellcolor[HTML]{FAF1D1} 1.96}          & {\color[HTML]{1F2329}\cellcolor[HTML]{FAF1D1} 34.03}          & {\color[HTML]{1F2329}\cellcolor[HTML]{FAF1D1} 67}           & {\color[HTML]{1F2329}\cellcolor[HTML]{FAF1D1} }          & {\color[HTML]{1F2329}\cellcolor[HTML]{FAF1D1} 38}               & {\color[HTML]{1F2329}\cellcolor[HTML]{FAF1D1} }          \\
{\color[HTML]{1F2329} }                                    & {\color[HTML]{1F2329} +DPad}                                            & {\color[HTML]{1F2329} 7.44}          & {\color[HTML]{1F2329} 12.07}          & {\color[HTML]{1F2329} 90}           & {\color[HTML]{1F2329} }          & {\color[HTML]{1F2329} 41.2}             & {\color[HTML]{1F2329} }       \\
{\color[HTML]{1F2329} }                                    & {\color[HTML]{1F2329} \cellcolor[HTML]{FAF1D1}+DPad+LaCache}                                       & {\color[HTML]{1F2329} \cellcolor[HTML]{FAF1D1}7.52}          & {\color[HTML]{1F2329} \cellcolor[HTML]{FAF1D1}12.23}          & {\color[HTML]{1F2329} \cellcolor[HTML]{FAF1D1}92}           & {\color[HTML]{1F2329}\cellcolor[HTML]{FAF1D1} }          & {\color[HTML]{1F2329} \cellcolor[HTML]{FAF1D1}38.4}             & {\color[HTML]{1F2329} \cellcolor[HTML]{FAF1D1}}       \\
{\color[HTML]{1F2329} }                                    & {\color[HTML]{1F2329} +Parallel+DPad}                                   & {\color[HTML]{1F2329} 2.21}          & {\color[HTML]{1F2329} 40.55}          & {\color[HTML]{1F2329} 90}           & {\color[HTML]{1F2329} }          & {\color[HTML]{1F2329} 40.2}             & {\color[HTML]{1F2329} }       \\
\multirow{-8}{*}{{\color[HTML]{1F2329} MBPP(3-shot)}}      & {\color[HTML]{1F2329} \cellcolor[HTML]{FAF1D1}\textbf{+Parallel+DPad+LaCache}}                     & {\color[HTML]{1F2329} \cellcolor[HTML]{FAF1D1}\textbf{2.26}} & {\color[HTML]{1F2329} \cellcolor[HTML]{FAF1D1}\textbf{40.01}} & {\color[HTML]{1F2329} \cellcolor[HTML]{FAF1D1}\textbf{91}}  & {\color[HTML]{1F2329} \cellcolor[HTML]{FAF1D1}\textbf{}} & {\color[HTML]{1F2329} \cellcolor[HTML]{FAF1D1}\textbf{40.2}}    & {\color[HTML]{1F2329} \cellcolor[HTML]{FAF1D1}\textbf{}}   \\
\bottomrule
\end{tabular}
\end{table*}

\begin{table}[t]
\centering
\caption{Ablation study on LLaDA-instruct under different context lengths.}
\label{tab:long-seq}
\begin{tabular}{lcc|cc}
\toprule
\multirow{2}{*}{Method} 
& \multicolumn{2}{c}{Length = 1024} 
& \multicolumn{2}{c}{Length = 2048} \\
\cmidrule(lr){2-3} \cmidrule(lr){4-5}
& HumanEval &  & HumanEval &  \\
\midrule
Vanilla                        & 37.8  &  & 37.8  &  \\
\cellcolor[HTML]{FAF1D1}{\color[HTML]{1F2329}+LaCache}                          & \cellcolor[HTML]{FAF1D1}{\textbf{40.85}} &  & \cellcolor[HTML]{FAF1D1}{39.63} &  \\
+Parallel (Fast-dLLM)          & 37.8  &  & 37.8  &  \\
\cellcolor[HTML]{FAF1D1}{+Parallel + LaCache}               & \cellcolor[HTML]{FAF1D1}{\textbf{41.46}} &  & \cellcolor[HTML]{FAF1D1}{39.63} &  \\
+DPad                          & 38.41 &  & \textbf{39.94} &  \\
\cellcolor[HTML]{FAF1D1}{+DPad + LaCache}                   & \cellcolor[HTML]{FAF1D1}{37.8}  &  & \cellcolor[HTML]{FAF1D1}{38.41} &  \\
+Parallel + DPad               & \textbf{41.46} &  & 39.33 &  \\
\cellcolor[HTML]{FAF1D1}{+Parallel + DPad + LaCache}        & \cellcolor[HTML]{FAF1D1}{40.85} &  & \cellcolor[HTML]{FAF1D1}{39.33} &  \\
\bottomrule
\end{tabular}
\end{table}

\begin{table*}[t]
\centering
\caption{Comprehensive benchmark results of Fast-dllm-V2 with FACache only.}

\label{tab:fastdllmv2-result}
\begin{tabular}{lllllll}
\cmidrule(lr){1-7} 
{\color[HTML]{333333} \textbf{Benchmark}}        & {\color[HTML]{333333} \textbf{Method}} & \multicolumn{3}{l}{{\color[HTML]{333333} \textbf{Efficiency}}}                                        & \multicolumn{2}{l}{{\color[HTML]{333333} \textbf{Accuracy(\%)}}} \\
\cmidrule(lr){3-5} \cmidrule(lr){6-7}  
{\color[HTML]{333333} }                          & {\color[HTML]{333333} }                & {\color[HTML]{333333} Latency(s)} & {\color[HTML]{333333} TPS}    & {\color[HTML]{333333} Gen.Length} & {\color[HTML]{333333} Flexible}                &                 \\
\cmidrule(lr){1-7} 
{\color[HTML]{333333} GSM8K(4-shot),bsz=256}     & {\color[HTML]{333333} Vanilla}         & {\color[HTML]{333333} 1.06}       & {\color[HTML]{333333} 310.19} & {\color[HTML]{333333} 328}        & {\color[HTML]{333333} 82.79}                   &                 \\

{\color[HTML]{333333} }                          & \cellcolor[HTML]{FAF1D1}{\color[HTML]{333333} +LaCache(FACache)}  & \cellcolor[HTML]{FAF1D1}{\color[HTML]{333333} 1}          & \cellcolor[HTML]{FAF1D1}{\color[HTML]{333333} 326.99} & \cellcolor[HTML]{FAF1D1}{\color[HTML]{333333} 328}        & \cellcolor[HTML]{FAF1D1}{\color[HTML]{333333} 82.79}                   &                 \\
\cmidrule(lr){1-7} 
{\color[HTML]{333333} MATH(4-shot),bsz=256}      & {\color[HTML]{333333} Vanilla}         & {\color[HTML]{333333} 1.31}       & {\color[HTML]{333333} 491.52} & {\color[HTML]{333333} 643}        & {\color[HTML]{333333} 59.94}                   &                 \\
{\color[HTML]{333333} }                          & \cellcolor[HTML]{FAF1D1}{\color[HTML]{333333} +LaCache(FACache)}  & \cellcolor[HTML]{FAF1D1}{\color[HTML]{333333} 1.23}       & \cellcolor[HTML]{FAF1D1}{\color[HTML]{333333} 523.48} & \cellcolor[HTML]{FAF1D1}{\color[HTML]{333333} 643}        & \cellcolor[HTML]{FAF1D1}{\color[HTML]{333333} 59.94}                   &                 \\
\cmidrule(lr){1-7} 
{\color[HTML]{333333} MBPP(3-shot),bsz=256}      & {\color[HTML]{333333} Vanilla}         & {\color[HTML]{333333} 0.9}        & {\color[HTML]{333333} 315.53} & {\color[HTML]{333333} 270}        & {\color[HTML]{333333} 31.2}                    &                 \\
{\color[HTML]{333333} }                          & \cellcolor[HTML]{FAF1D1}{\color[HTML]{333333} +LaCache(FACache)}  & \cellcolor[HTML]{FAF1D1}{\color[HTML]{333333} 0.84}       & \cellcolor[HTML]{FAF1D1}{\color[HTML]{333333} 336.92} & \cellcolor[HTML]{FAF1D1}{\color[HTML]{333333} 270}        & \cellcolor[HTML]{FAF1D1}{\color[HTML]{333333} 31.2}                    &                 \\
\cmidrule(lr){1-7} 
{\color[HTML]{333333} HumanEval(0-shot),bsz=256} & {\color[HTML]{333333} Vanilla}         & {\color[HTML]{333333} 1.57}       & {\color[HTML]{333333} 269.68} & {\color[HTML]{333333} 369}        & {\color[HTML]{333333} 65.24}                   &                 \\
{\color[HTML]{333333} }                          & \cellcolor[HTML]{FAF1D1}{\color[HTML]{333333} +LaCache(FACache)}  & \cellcolor[HTML]{FAF1D1}{\color[HTML]{333333} 1.45}       & \cellcolor[HTML]{FAF1D1}{\color[HTML]{333333} 289.59} & \cellcolor[HTML]{FAF1D1}{\color[HTML]{333333} 362}        & \cellcolor[HTML]{FAF1D1}{\color[HTML]{333333} 65.24}                   &  \\
\cmidrule(lr){1-7}  \\
\end{tabular}
\end{table*}

\begin{table*}[t]
\centering
\caption{Comprehensive results On an another designed GPU of LlaDA-Instruct with LaCache.}
\label{tab:more-GPU}
\footnotesize
\begin{tabular}{ccccccc}
\cmidrule(lr){1-7} 
{\color[HTML]{1F2329} }                                                       & 

{\color[HTML]{1F2329} }                              & \multicolumn{3}{c}{{\color[HTML]{1F2329} Efficiency}}                                                                                              & \multicolumn{2}{c}{{\color[HTML]{1F2329} Accuracy(\%)}}                                                     \\
\cmidrule(lr){3-5} \cmidrule(lr){6-7}  \\
\multirow{-2}{*}{{\color[HTML]{1F2329} Benchmark}}                            & \multirow{-2}{*}{{\color[HTML]{1F2329} Method}}      & {\color[HTML]{1F2329} Latency(s)}                   & {\color[HTML]{1F2329} TPS}                           & {\color[HTML]{1F2329} Gen.Length}                  & {\color[HTML]{1F2329} Flexible}                      & {\color[HTML]{1F2329} Strict/acc\_norm}              \\
\cmidrule(lr){1-7} 
{\color[HTML]{1F2329} }                                                       & {\color[HTML]{1F2329} Vanilla}                       & {\color[HTML]{1F2329} 11.07}                        & {\color[HTML]{1F2329} 20.82}                         & {\color[HTML]{1F2329} 230}                         & {\color[HTML]{1F2329} 76.04}                         & {\color[HTML]{1F2329} 35.41}                         \\
\multirow{-2}{*}{{\color[HTML]{1F2329} GSM8K(4-shot)}}                        & \cellcolor[HTML]{FAF1D1}{\color[HTML]{1F2329} +LaCache} & \cellcolor[HTML]{FAF1D1}{\color[HTML]{1F2329} 8.41} & \cellcolor[HTML]{FAF1D1}{\color[HTML]{1F2329} 27.5}  & \cellcolor[HTML]{FAF1D1}{\color[HTML]{1F2329} 231} & \cellcolor[HTML]{FAF1D1}{\color[HTML]{1F2329} 75.66} & \cellcolor[HTML]{FAF1D1}{\color[HTML]{1F2329} 35.56} \\
\cmidrule(lr){1-7} 
{\color[HTML]{1F2329} }                                                       & {\color[HTML]{1F2329} Vanilla}                       & \cellcolor[HTML]{FFFFFF}{\color[HTML]{1F2329} 9.58} & \cellcolor[HTML]{FFFFFF}{\color[HTML]{1F2329} 25.96} & \cellcolor[HTML]{FFFFFF}{\color[HTML]{1F2329} 249} & {\color[HTML]{1F2329} 8.5}                           & {\color[HTML]{1F2329} 38.5}                          \\
\multirow{-2}{*}{{\color[HTML]{1F2329} MATH(4-shot)}}                         & \cellcolor[HTML]{FAF1D1}{\color[HTML]{1F2329} +LaCache} & \cellcolor[HTML]{FAF1D1}{\color[HTML]{1F2329} 7.93} & \cellcolor[HTML]{FAF1D1}{\color[HTML]{1F2329} 31.32} & \cellcolor[HTML]{FAF1D1}{\color[HTML]{1F2329} 248} & \cellcolor[HTML]{FAF1D1}{\color[HTML]{1F2329} 7.88}  & \cellcolor[HTML]{FAF1D1}{\color[HTML]{1F2329} 38.82} \\
\cmidrule(lr){1-7} 
\rowcolor[HTML]{FFFFFF} 
\cellcolor[HTML]{FFFFFF}{\color[HTML]{1F2329} }                               & {\color[HTML]{1F2329} Vanilla}                       & {\color[HTML]{1F2329} 25.1}                         & {\color[HTML]{1F2329} 11.91}                         & {\color[HTML]{1F2329} 299}                         & {\color[HTML]{1F2329} 15.58}                         & {\color[HTML]{1F2329} }                              \\
\rowcolor[HTML]{EEF6C6} 
\multirow{-2}{*}{\cellcolor[HTML]{FFFFFF}{\color[HTML]{1F2329} MBPP(3-shot)}} & \cellcolor[HTML]{FAF1D1}{\color[HTML]{1F2329} +LaCache}                         & \cellcolor[HTML]{FAF1D1}{\color[HTML]{1F2329} 18.01}                        & \cellcolor[HTML]{FAF1D1}{\color[HTML]{1F2329} 16.64}                         & \cellcolor[HTML]{FAF1D1}{\color[HTML]{1F2329} 300}                         & \cellcolor[HTML]{FAF1D1}{\color[HTML]{1F2329} 14.8}                          & \cellcolor[HTML]{FAF1D1}{\color[HTML]{1F2329} }                              \\
\cmidrule(lr){1-7} 
{\color[HTML]{1F2329} }                                                       & {\color[HTML]{1F2329} Vanilla}                       & {\color[HTML]{1F2329} -}    & {\color[HTML]{1F2329} -}     & {\color[HTML]{1F2329} -}   & {\color[HTML]{1F2329} 78.7}                          & {\color[HTML]{1F2329} 76.6}                          \\
\multirow{-2}{*}{{\color[HTML]{1F2329} Arc\_Easy}}                            & \cellcolor[HTML]{FAF1D1}{\color[HTML]{1F2329} +LaCache} & {\color[HTML]{1F2329} -}    & {\color[HTML]{1F2329} -}     & {\color[HTML]{1F2329} -}   & \cellcolor[HTML]{FAF1D1}{\color[HTML]{1F2329} 79.46} & \cellcolor[HTML]{FAF1D1}{\color[HTML]{1F2329} 76.18} \\

\cmidrule(lr){1-7} 
{\color[HTML]{1F2329} }                                                       & {\color[HTML]{1F2329} Vanilla}                       & {\color[HTML]{1F2329} -}    & {\color[HTML]{1F2329} -}     & {\color[HTML]{1F2329} -}   & {\color[HTML]{1F2329} 50}                            & {\color[HTML]{1F2329} 54.5}                          \\
\multirow{-2}{*}{{\color[HTML]{1F2329} Arc\_Chanllenge}}                      & \cellcolor[HTML]{FAF1D1}{\color[HTML]{1F2329} +LaCache} & {\color[HTML]{1F2329} -}    & {\color[HTML]{1F2329} -}     & {\color[HTML]{1F2329} -}   & \cellcolor[HTML]{FAF1D1}{\color[HTML]{1F2329} 51.28} & \cellcolor[HTML]{FAF1D1}{\color[HTML]{1F2329} 52.99} \\
\cmidrule(lr){1-7} 
{\color[HTML]{1F2329} }                                                       & {\color[HTML]{1F2329} Vanilla}                       & {\color[HTML]{1F2329} -}    & {\color[HTML]{1F2329} -}     & {\color[HTML]{1F2329} -}   & {\color[HTML]{1F2329} 69.53}                         & {\color[HTML]{1F2329} -}                             \\
\multirow{-2}{*}{{\color[HTML]{1F2329} Winogrande}}                           & \cellcolor[HTML]{FAF1D1}{\color[HTML]{1F2329} +LaCache} & {\color[HTML]{1F2329} -}    & {\color[HTML]{1F2329} -}     & {\color[HTML]{1F2329} -}   & \cellcolor[HTML]{FAF1D1}{\color[HTML]{1F2329} 69.06} & \cellcolor[HTML]{FAF1D1}{\color[HTML]{1F2329} -}     \\
\cmidrule(lr){1-7} 
{\color[HTML]{1F2329} }                                                       & {\color[HTML]{1F2329} Vanilla}                       & {\color[HTML]{1F2329} -}    & {\color[HTML]{1F2329} -}     & {\color[HTML]{1F2329} -}   & \cellcolor[HTML]{FAF1D1}{\color[HTML]{1F2329} 28.35} & \cellcolor[HTML]{FAF1D1}{\color[HTML]{1F2329} 28.35} \\
\multirow{-2}{*}{{\color[HTML]{1F2329} GPQA}}                                 & \cellcolor[HTML]{FAF1D1}{\color[HTML]{1F2329} +LaCache} & {\color[HTML]{1F2329} -}    & {\color[HTML]{1F2329} -}     & {\color[HTML]{1F2329} -}   & \cellcolor[HTML]{FAF1D1}{\color[HTML]{1F2329} 30.13} & \cellcolor[HTML]{FAF1D1}{\color[HTML]{1F2329} 30.13} \\
\cmidrule(lr){1-7} 
\end{tabular}
\end{table*}

\begin{algorithm*}[]
\caption{the first layer's Flash-Attention}
\begin{algorithmic}[1]
\Require Matrices $Q, K, V, Ocache \in \mathbb{R}^{N \times d}, \ell{cache}, RowMaxcache \in \mathbb{R}^{N \times 4d}$ in HBM, block sizes $B_c, B_r$, $GenStep \in \{True,False\}$, where $GenStep=True$ if it's in the first generation step.
\State Divide $Q$ into $T_r = \lceil N / B_r \rceil$ blocks $Q_1, \ldots, Q_{T_r}$ of size $B_r \times d$, divide $K, V$ into $T_c = \lceil N / B_c \rceil$ blocks $K_1, \ldots, K_{T_c}$ and $V_1, \ldots, V_{T_c}$ of size $B_c \times d$, divide the cache $Ocache \in \mathbb{R}^{N \times d}$ into $T_r$ blocks $Ocache_1, \ldots, Ocache_{T_r}$ of size $B_r \times d$, divide the cache $\ell{cache}, RowMaxcache \in \mathbb{R}^{N \times 4d}$ into $T_r$ blocks $\ell{cache}_1, \ldots, \ell{cache}_{T_r}$ and $RowMaxcache_1, \ldots, RowMaxcache_{T_r}$ of size $B_r \times 4d$.
\State Divide the output $O \in \mathbb{R}^{N \times d}$ into $T_r$ blocks $O_1, \ldots, O_{T_r}$ of size $B_r \times d$, and divide the logsumexp $L$ into $T_r$ blocks $L_1, \ldots, L_{T_r}$ of size $B_r$. 

\State Define overlap indicator
\[
Q_{overlap} =
\begin{cases}
True, & \text{if } Q_i \text{ overlaps with current Block } \\
False, & \text{otherwise}.
\end{cases}\] \\ \[
K_{overlap} =
\begin{cases}
True, & \text{if } K_j \text{ overlaps with current Block } \\
False, & \text{otherwise}.
\end{cases}
\]
\For{$1 \le i \le T_r$}
    \State Load $Q_i$ from HBM to on-chip SRAM.
    \State On chip, initialize
        $O_i^{(0)} = \mathbf{0}_{B_r \times d}, \quad
        \ell_i^{(0)} = \mathbf{0}_{B_r}, \quad
        m_i^{(0)} = (-\infty)_{B_r}$.
    \If{$GenStep=True$ \textbf{and} $Q_{overlap}=False$}
     \State
        $Ocache_i^{(0)} = \mathbf{0}_{B_r \times d},  \quad \ell{cache}_i^{(0)} = \mathbf{0}_{B_r \times 4d},  \quad RowMaxcache_i^{(0)} = \mathbf{0}_{B_r \times 4d}$.
     \ElsIf{$GenStep=False$ \textbf{and} $Q_{overlap}=False$}
        \State $O_i^{(0)} = Ocache_i^{(0)}, \quad
        \ell_i^{(0)} = \ell{cache}_i^{(0)}, \quad
        m_i^{(0)} = RowMaxcache_i^{(0)}$. \Comment{Initialize with cache}
     \EndIf
    \For{$1 \le j \le T_c$}
        \If{$GenStep=Flase $ \textbf{and} $ Q_{overlap}=False $ \textbf{and} $ K_{overlap}=False $} 
                \State \textbf{continue}.\Comment{Already calculated, skip}
        \EndIf
        \State Load $K_j, V_j$ from HBM to on-chip SRAM.
        \State On chip, compute $S_i^{(j)} = Q_i K_j^\top \in \mathbb{R}^{B_r \times B_c}.$
        \State On chip, compute $m_i^{(j)} = \max\!\left(m_i^{(j-1)}, \operatorname{rowmax}(S_i^{(j)})\right),
            \tilde{P}_i^{(j)} = \exp\!\left(S_i^{(j)} - m_i^{(j)}\right) \in \mathbb{R}^{B_r \times B_c},$ 
            \State $\ell_i^{(j)} = e^{m_i^{(j-1)} - m_i^{(j)}} \ell_i^{(j-1)} + \operatorname{rowsum}(\tilde{P}_i^{(j)}).$
        \State On chip, compute $O_i^{(j)} =
            \operatorname{diag}\!\left(e^{m_i^{(j-1)} - m_i^{(j)}}\right) O_i^{(j-1)}
            + \tilde{P}_i^{(j)} V_j.$
        \If{$GenStep=True $ \textbf{and} $ Q_{overlap}=False $ \textbf{and} $ K_{overlap}=False $} 
        \State $Ocache_i^{(j)} =
            \operatorname{diag}\!\left(e^{m_i^{(j-1)} - m_i^{(j)}}\right) Ocache_i^{(j-1)}
            + \tilde{P}_i^{(j)} V_j,$
        \State $\ell{cache}_i = e^{m_i^{(j-1)} - m_i^{(j)}}\ell{cache}_i + \operatorname{rowsum}(\tilde{P}_i^{(j)}).$ \Comment{Update cache}
        \EndIf
    \EndFor
    \State On chip, compute $O_i = \operatorname{diag}\!\left(\ell_i^{(T_c)}\right)^{-1} O_i^{(T_c)}.$
    \State On chip, compute$L_i = m_i^{(T_c)} + \log\!\left(\ell_i^{(T_c)}\right).$
    \State Write $O_i$ to HBM as the $i$-th block of $O$.
    \State Write $L_i$ to HBM as the $i$-th block of $L$.
    \If{$GenStep=True $ \textbf{and} $ Q_{overlap}=False$} 
        \State Write $Ocache_i$ to HBM as the $i$-th block of $Ocache$.
        \State Write $\ell{cache}_i$ to HBM as the $i$-th block of $\ell{cache}$.
        \State Write $RowMaxcache_i$ to HBM as the $i$-th block of $RowMaxcache$. \Comment{Store cache}
    \EndIf
\EndFor
\end{algorithmic}
\end{algorithm*}

\end{document}